\documentclass{article}

\PassOptionsToPackage{sort&compress}{natbib}
\usepackage[preprint]{corl_2026} 

\usepackage{booktabs}
\usepackage{adjustbox}
\usepackage{xspace}

\usepackage{algorithm}
\usepackage[noend]{algpseudocode}
\usepackage{amssymb}
\usepackage[normalem]{ulem}
\usepackage[font=small,labelfont=bf]{caption}

\usepackage[utf8]{inputenc} 
\usepackage[T1]{fontenc}    
\usepackage{hyperref}       
\usepackage{url}            
\usepackage{booktabs}       
\usepackage{amsfonts}       
\usepackage{nicefrac}       
\usepackage{microtype}      
\usepackage{wrapfig}
\usepackage{pifont}

\usepackage{comment,url,graphicx,subcaption,relsize} 
\usepackage{amssymb,amsfonts,amsmath,amsthm,amscd,dsfont,mathrsfs,mathtools,nicefrac}
\usepackage{float,psfrag,epsfig,color,soul,url,hyperref}
\usepackage[font=small,labelfont=bf]{caption}
\usepackage{epstopdf,bbm,mathtools,enumitem}
\usepackage{stackrel}
\usepackage[table,xcdraw]{xcolor}

\newcommand{\solstart}[1] {\vspace{.25in} \textbf{Solution to Q#1} \vspace{.10in}}

\newcommand{\solend}{\newpage\setcounter{page}{1}\setcounter{equation}{0}}

\def\balign#1\ealign{\begin{align}#1\end{align}}
\def\baligns#1\ealigns{\begin{align*}#1\end{align*}}
\def\balignat#1\ealign{\begin{alignat}#1\end{alignat}}
\def\balignats#1\ealigns{\begin{alignat*}#1\end{alignat*}}
\def\bitemize#1\eitemize{\begin{itemize}#1\end{itemize}}
\def\benumerate#1\eenumerate{\begin{enumerate}#1\end{enumerate}}

\newenvironment{talign*}
 {\csname align*\endcsname}
 {\endalign}
\newenvironment{talign}
 {\csname align\endcsname}
 {\endalign}

\def\balignst#1\ealignst{\begin{talign*}#1\end{talign*}}
\def\balignt#1\ealignt{\begin{talign}#1\end{talign}}

\let\originalleft\left
\let\originalright\right
\renewcommand{\left}{\mathopen{}\mathclose\bgroup\originalleft}
\renewcommand{\right}{\aftergroup\egroup\originalright}

\def\tinycitep*#1{{\tiny\citep*{#1}}}
\def\tinycitealt*#1{{\tiny\citealt*{#1}}}
\def\tinycite*#1{{\tiny\cite*{#1}}}
\def\smallcitep*#1{{\scriptsize\citep*{#1}}}
\def\smallcitealt*#1{{\scriptsize\citealt*{#1}}}
\def\smallcite*#1{{\scriptsize\cite*{#1}}}

\def\mbb#1{\mathbb{#1}}

\def\<{\left\langle} 
\def\>{\right\rangle}

\def\E{\mbb{E}} 

\def\Esubarg#1#2{\E_{#1}\left[{#2}\right]}

\ifdefined\nonewproofenvironments\else
\ifdefined\ispres\else

\newenvironment{proof-sketch}{\noindent\textbf{Proof Sketch}
  \hspace*{1em}}{\qed\bigskip\\}
\newenvironment{proof-idea}{\noindent\textbf{Proof Idea}
  \hspace*{1em}}{\qed\bigskip\\}
\newenvironment{proof-of-lemma}[1][{}]{\noindent\textbf{Proof of Lemma {#1}}
  \hspace*{1em}}{\qed\\}
\newenvironment{proof-of-theorem}[1][{}]{\noindent\textbf{Proof of Theorem {#1}}
  \hspace*{1em}}{\qed\\}
\newenvironment{proof-attempt}{\noindent\textbf{Proof Attempt}
  \hspace*{1em}}{\qed\bigskip\\}

\fi

\fi

\usepackage{titlesec}
\titlespacing{\section}{0pt}{0.3\baselineskip}{0.25\baselineskip}
\titlespacing{\subsection}{0pt}{0.25\baselineskip}{0.15\baselineskip}
\titlespacing{\subsubsection}{0pt}{0.05\baselineskip}{0.03\baselineskip}

\renewcommand{\paragraph}[1]{\vspace{0.2em}\noindent\textit{#1} --}

\newcommand{\spire}{\textsc{SPIRE}}

\newcommand{\AlgoName}{ReinforceGen}

\newif\ifcomments
\commentsfalse

\ifcomments
    \newcommand{\reminder}[1]{\textcolor{red}{(Reminder: #1)}}
    \newcommand{\todo}[1]{\textcolor{red}{(TODO: #1)}}
    \newcommand{\caelan}[1]{\textcolor{blue}{(CG: #1)}}
    \newcommand{\ajay}[1]{\textcolor{green}{(Ajay: #1)}}
    \newcommand{\zihan}[1]{\textcolor{purple}{(Zihan: #1)}}
\else
    \newcommand{\reminder}[1]{}
    \newcommand{\todo}[1]{}
    \newcommand{\caelan}[1]{}
    \newcommand{\ajay}[1]{}
    \newcommand{\zihan}[1]{}
\fi

\newif\ifcommentsnew
\commentsnewfalse

\ifcommentsnew
    \newcommand{\new}[1]{\textcolor{orange}{(Caelan: #1)}}
    \newcommand{\zihannew}[1]{\textcolor{purple}{(Zihan: #1)}}
\else
    \newcommand{\new}[1]{}
    \newcommand{\zihannew}[1]{}
\fi

\newcommand{\rebuttal}[2]{#2}

\title{ReinforceGen: Hybrid Skill Policies with Automated Data Generation and Reinforcement Learning}

\author{
  Zihan Zhou$^{1}$,
  Animesh Garg$^{2}$,
  Ajay Mandlekar$^3$$^*$,
  Caelan Garrett$^3$$^*$, \\
  $^1$ University of Toronto, Vector Institute $^2$ Georgia Institute of Technology $^3$ NVIDIA Research \\ 
  $^*$ equal advising
}

\begin{document}
\maketitle

\begin{abstract}
Long-horizon manipulation has been a long-standing challenge in the robotics community. We propose \AlgoName{}, a system that combines task decomposition, data generation, imitation learning, and motion planning to form an initial solution, and improves each component through reinforcement-learning-based fine-tuning. \AlgoName{} first segments the task into multiple localized skills, which are connected through motion planning. The skills and motion planning targets are trained with imitation learning on a dataset generated from 10 human demonstrations, and then fine-tuned through online adaptation and reinforcement learning. When benchmarked on the Robosuite dataset, \AlgoName{} reaches 80\% success rate on all tasks with visuomotor controls in the highest reset range setting. Additional ablation studies show that our fine-tuning approaches contribute to an 89\% average performance increase. Finally, \AlgoName{} demonstrates significant improvement through fine-tuning in our real-world evaluations. More results and videos are available at \href{https://reinforcegen.github.io}{\texttt{reinforcegen.github.io}}.

\end{abstract}


\section{Introduction}\label{sec:intro}

Imitation Learning (IL) from demonstrations is an effective approach for robots to autonomously complete tasks. 
In long-horizon tasks, collecting demonstrations can be expensive, and the trained agent is more likely to deviate from the demonstrations to out-of-distribution states. Reinforcement Learning (RL) leverages random exploration, 
incorporating environmental feedback through rewards. However, long horizons exacerbate the exploration challenge, especially when the reward signals are also sparse.
In the context of robot learning, collecting demonstrations for IL is often time-consuming and expensive, as it typically requires a teleoperation platform and coordinating with human operators. Furthermore, the solution quality and data coverage of the demonstrations are critical, as they directly impact the agent's performance when used in methods such as Behavior Cloning (BC)~\citep{mandlekar2021matters}. 

One approach to combat demonstration insufficiency is to augment the dataset through synthetic data generation. In robotic manipulation tasks, a line of work~\citep{mimicgen,garrett2024skillmimicgen,jiang2024dexmimicen} focuses on object-centric data generation through demonstration adaptation. Other approaches~\citep{mcdonald2022guided,dalal2023imitating} use Task and Motion Planning (TAMP)~\citep{garrett2021integrated} to generate demonstrations.
An alternative strategy is to hierarchically divide the task into consecutive stages with easier-to-solve subgoals~\citep{mandlekar2023hitltamp,garrett2024skillmimicgen,zhou2024spire}. 
In most manipulation tasks, only a small fraction of robot execution
requires high-precision movements, for example, only the contact-rich segments.
Thus, these approaches concentrate the demo collection effort at the precision-intensive skill segments and connect segments using planning, ultimately improving demo sample efficiency.

Still, these demonstration generation methods are open-loop and rely solely on offline data. 
As a result, IL agents trained with the generated data are still bottlenecked by the quality of the source demonstrations.  
To combat this, we propose \AlgoName{}, a framework that improves hierarchical data generation by incorporating online exploration and environmental feedback using RL.
\AlgoName{} trains a hybrid BC agent with object-centric data generation as its base policy. It then combines distillation, causal inference, and RL to improve the base agent with online data, as well as real-time adaptation from environment feedback during deployment.
We demonstrate that \AlgoName{} produces high-performance hybrid data generators 
and show that \AlgoName{}-generated data can also be used to train end-to-end visuomotor imitation policies.
We also demonstrate that \AlgoName{} can be effectively deployed on a real robot through zero-shot sim-to-real transfer.

\begin{figure}[t]
    \centering
    \includegraphics[width=1.0\linewidth]{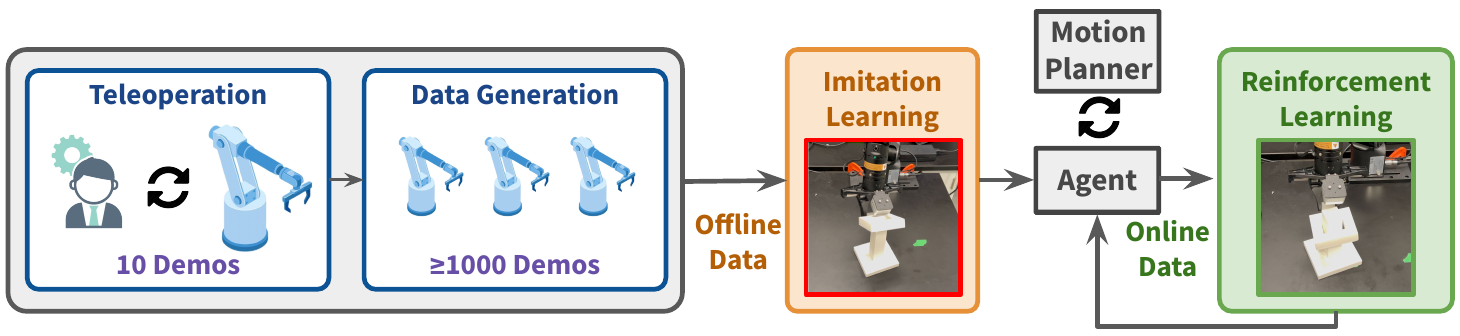}
    \caption{\AlgoName{} first creates an offline dataset by synthetic data generation from a small set of source human demonstrations. The dataset is then used to train a hybrid imitation learning agent that alternates between moving to a predicted waypoint using a motion planner and directly controlling the robot using a learned policy. Finally, \AlgoName{} uses reinforcement learning to fine-tune the agent with online environment interactions.}
    \label{fig:enter-label}
\end{figure}

\textbf{The contributions of this paper are the following.}
\newline \noindent $\bullet$ We propose \AlgoName{}, an automated demonstration generation system that integrates planning, behavior cloning, and reinforcement learning to train policies that robustly solve long-horizon and contact-rich tasks using only a handful of human demonstrations.
\vspace{1mm}
\newline \noindent $\bullet$ Through localized reinforcement learning, \AlgoName{} is able to explore and thus go beyond existing automated demonstration systems, which are bounded by the performance of the demonstrator.
\vspace{1mm}
\newline \noindent $\bullet$ We evaluate \AlgoName{} on multi-stage contact-rich manipulation tasks. \AlgoName{} reaches an \textbf{80\%} success rate, almost doubling the success rate of the prior state-of-the-art~\citep{garrett2024skillmimicgen}.
\vspace{1mm}
\newline \noindent $\bullet$ Finally, we deploy \AlgoName{} on a real robot through zero-shot sim-to-real transfer. \AlgoName{} agent achieves a \textbf{70\%} success rate, a $2.8\times$ improvement over prior state-of-the-art.
\section{Related Work}\label{sec:related-work}

\textbf{Imitation and reinforcement learning for robotics.} IL from human demonstrations has been a pillar in robot learning, benchmarking significant results in both simulations and real-world applications~\citep{florence2022implicit, diffusionpolicy, bet, act, kim24openvla}. RL is another promising approach to solving robotic problems with even real-world successes~\citep{annurev:/content/journals/10.1146/annurev-control-030323-022510, Wu2022DayDreamerWM, yu2023language}. However, without engineered rewards, RL often struggles with exploration and can even outright fail~\citep{zhou2024spire}. Similar to our approach, ~\cite{Nair2017OvercomingEI, Johannink2018ResidualRL, zhou2024spire} investigate combining the strengths of IL and RL, using the IL policy as a warm start for RL exploration and using RL to fine-tune the IL agent in out-of-distribution states.
However, \AlgoName{} also leverages motion planning to decompose control into smaller skill-learning subproblems, which decreases the horizon required for RL to explore and thus the state space, dramatically boosting policy success rates.

\textbf{Automated data generation.} \cite{mcdonald2022guided,dalal2023imitating} use Task and Motion Planning (TAMP)~\citep{garrett2021integrated} to generate demonstrations for imitation learning; however, these systems focus on prehensile manipulation due to the difficulty of modeling contact-rich skills. 
\cite{dalal2024psl} leverages Large Language Models (LLMs) for task planning in place of a full planning model. 
MimicGen~\citep{mimicgen}, SkillMimicGen~\citep{garrett2024skillmimicgen}, DexMimicGen~\citep{jiang2024dexmimicen}, 
DemoGen\citep{xue2025demogen}, CP-Gen~\citep{lin2025constraint}, and MoMaGen~\citep{li2025momagen} automatically bootstrap a dataset of demonstrations from a few annotated human source demonstrations through pose adaptation and trial-and-error execution.
A key drawback is that the generated dataset has limited diversity 
because each demonstration is derived from the same small set.
In \AlgoName{}, we address this by using reinforcement learning to explore beyond these demonstrations in search of more successful and less costly behaviors.

\textbf{Hybrid planning and learning systems.} Several approaches integrate learned policies into TAMP systems.
\cite{wang2021learning} engineered open-loop parameterized skill policies and learned successful skill parameters using Gaussian Process regression.
HITL-TAMP~\citep{mandlekar2023hitltamp} learned closed-loop visuomotor policies in place of open-loop skills using BC.
The closest to our approach is SPIRE~\citep{zhou2024spire}, which goes beyond HITL-TAMP by using RL to fine-tune initiation-learned skills, improving their success rates and decreasing execution costs.
In order to leverage TAMP, all these approaches assume access to an engineering planning model that specifies the preconditions and effects of each skill.
In contrast, \AlgoName{} learns and fine-tunes initiation and termination models that implicitly encode these dynamics without prior information.

\section{Preliminaries}\label{sec:method}

\begin{figure}
    \centering
    \includegraphics[width=0.8\linewidth]{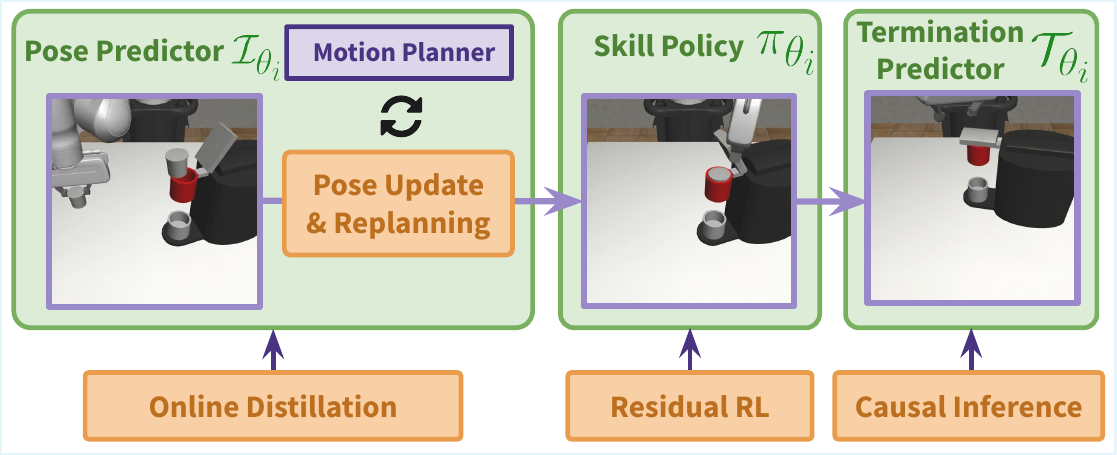}
    \caption{The three main components of a \AlgoName{} stage. The pose predictor $\mathcal{I}_{\theta_i}$ predicts the target end-effector pose and updates the goal of the motion planner in real-time. After reaching the destination, the skill policy $\pi_{\theta_i}$ takes control to complete the stage goal, which is determined by the termination predictor $\mathcal{T}_{\theta_i}$. All three components are first trained offline with supervised learning and then fine-tuned with online data.}
    \label{fig:enter-label}
\end{figure}

We model manipulation tasks as Partially Observable Markov Decision Processes (POMDPs)
where $\mathcal{S}$ is the state space, $\mathcal{O}$ is the observation space, $\mathcal{A}$ is the action space, and $\mathcal{T}\subseteq\mathcal{S}$ 
is a set of terminal states. 
We are interested in producing a policy $\pi:\mathcal{O}\rightarrow\mathcal{A}$ that controls the system from initial state $s_0 \in \mathcal{S}$ to a terminal state $s_T \in \mathcal{T}$ while minimizing the number of steps taken.

\subsection{Object-Centric Data Generation} 
\label{sec:data-gen}

\AlgoName{} builds on MimicGen~\citep{mimicgen}, which automatically generates data by transforming and replaying human demonstrations for manipulation tasks. 
It assumes the environment contains a set of manipulable objects $M=\{O_1,\dots, O_m\}$, and their poses comprise a component of states $s \in \mathcal{S}$.
MimicGen divides the task into multiple contiguous object-centric subtask segments, where each segment $i$ is associated with a fixed reference object $R_i\in M$. 
Let $T^A_B \in \text{SE}(3)$ be the homogeneous transformation matrix representing object $A$'s pose in the frame of object $B$, where $A, B \in M$ and $W$ is the world frame. 
For a demonstration involving object $A$, the source trajectories of end-effector poses $\{T_W^{A_t}\}$ are parsed into the frame of the corresponding reference objects $T_{R_i}^{A_t}\leftarrow (T_W^{R_i})^{-1}T_{W}^{A_t}$. To generate new data in a new environment instantiation, segments of a sampled source trajectory are
adapted according to the current reference object $R'_i$ pose $T_W^{A'_t}\leftarrow T_W^{R'_i}T_{R_i}^{A_t}$. 
Delta pose actions are extracted from the transformed poses and executed open loop along with the original gripper actions.
After execution, trajectories that reach a goal state comprise the new synthetic dataset.

\subsection{Hybrid Skill Policy}
\label{sec:hsp}

\AlgoName{} adopts the bilevel decomposition from the Hybrid Skill Policy (HSP) framework~\citep{garrett2024skillmimicgen} to reduce the complexity of long-horizon tasks by framing control as a sequence of connected contact-rich subproblems.
More formally, an HSP decomposes the task into $n$ stages, where each stage comprises a connect segment followed by a skill segment. In the connect segment, the robot moves from its 
current position to the initiation position of the skill segment.
Following the \emph{options} framework~\citep{Stolle2002LearningOI}, the $i$-th skill in the skill sequence $\psi_i:=\langle\mathcal{I}_i, \pi_i, \mathcal{T}_i\rangle$ consists of an initiation end-effector pose condition $\mathcal{I}_i\subseteq\text{SE}(3)$, a skill policy $\pi_i:\mathcal{O}\rightarrow\mathcal{A}$, and a termination condition $\mathcal{T}_i\subseteq\mathcal{S}$.
When applying this formation to object-centric data generation (Sec.~\ref{sec:data-gen}), we assume that the skill segment in each stage uses a reference object $R_i\in M$.
Initiation end-effector poses can also be generated in an object-centric fashion by transforming the target pose $E_0$ in the source trajectory in the frame of the original reference object to the current: $T^{E'_0}_W\leftarrow T^{R'_i}_WT^{E_0}_{R_i}$, as in \cite{garrett2024skillmimicgen}. 

States are not directly observable at deployment time, so we seek to learn parametric models of skills that operate only on observations.
Formally, an HSP \emph{agent} is a sequence of parameterized skills $[\psi_{\theta_1}, ..., \psi_{\theta_n}]$. 
Each skill $\psi_{\theta_i}:=\langle \mathcal{I}_{\theta_i}, \pi_{\theta_i}, \mathcal{T}_{\theta_i}\rangle$ comprises an {\em initiation predictor} $\mathcal{I}_{\theta_i}:\mathcal{O}\rightarrow\text{SE}(3)$ that predicts a skill start pose in $\mathcal{I}_i$ based on the current observation, a parameterized {\em skill policy} $\pi_{\theta_i}:\mathcal{O}\rightarrow\mathcal{A}$, and a {\em termination classifier} $\mathcal{T}_{\theta_i}:\mathcal{O}\rightarrow\{0, 1\}$ that predicts whether the current state is a terminal state in $\mathcal{T}_i$. 
The parameters $\{\theta_i\}$ are learned through behavior cloning, maximizing the likelihood of producing the generated data: $P(E'_0=\mathcal{I}_{\theta_i}(o_0))$, $P(a'_t=\pi_{\theta_i}(o_t))$, and $P([s_t\in\mathcal{T}_i]=\mathcal{T}_{\theta_i}(o_t))$.
At each stage, an HSP predicts a starting pose with $\mathcal{I}_{\theta_i}(o_0)$, moves to the pose with motion planning, and executes the skill policy $\pi_{\theta_i}$ until $\mathcal{T}_{\theta_i}$ predicts termination.
As in prior work~\citep{mimicgen, garrett2024skillmimicgen}, we assume the stage sequence and reference objects $R_i$ are annotated by a human.

\section{\AlgoName{}}

Existing object-centric data generation methods~\citep{mimicgen,garrett2024skillmimicgen, mandlekar2023hitltamp} have been shown to be capable of generating high-quality trajectory datasets that produce performant policies in long-horizon manipulation tasks from a small set of human demonstrations. 
Despite their impressive performance, these methods are still bounded by the quality of the demonstrations since they primarily replay transformed source trajectories or directly run IL on the collected demonstrations, without exploring new behaviors. \
\AlgoName{} presents a solution to improve data generation quality substantially by integrating a trained HSP imitation learning policy (Sec.~\ref{sec:hsp}) and online environment feedback into the data generation workflow, allowing the policy, and ultimately the generated data, to improve over time via exploration and exploitation. Specifically, we propose fine-tuning pipelines for individual parameterized components of an HSP with online adaptation and reinforcement learning, namely the
initiation pose predictor $\mathcal{I}_{\theta_i}$ (Sec.~\ref{sec:alg:init-cond}), skill policy $\pi_{\theta_i}$ (Sec.~\ref{sec:alg:policy}), and termination predictor $\mathcal{T}_{\theta_i}$ (Sec.~\ref{sec:alg:term-cond}).
With the fine-tuned components, we execute these skill policies in a hybrid policy that alternates between motion planning and the skills (Sec.~\ref{sec:alg:deployment}). 
Finally, we demonstrate the ability to optionally distill the hybrid policy to a fully end-to-end policy that does not rely on a motion planner (Sec.~\ref{sec:alg:distill}).

\subsection{Initiation Pose Prediction}
\label{sec:alg:init-cond}

\begin{wrapfigure}{r}{0.3\textwidth}
\vspace{-35px}

    \centering
    \includegraphics[width=\linewidth]{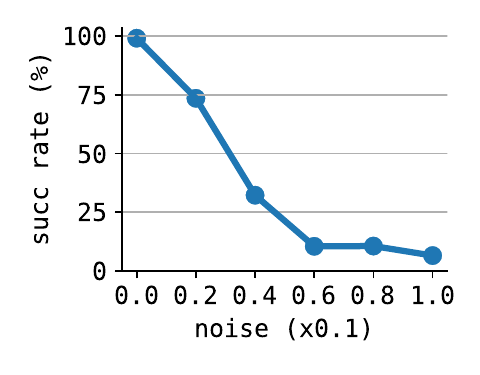}
    \caption{The policy success rate drops sharply as the pose target noise increases in the second stage of \emph{Nut Assembly}. \rebuttal{}{See App.~\ref{app:details:pose-noise} for more details.}}
    \label{fig:error-succ}
\vspace{-15px}
\end{wrapfigure}

The initiation pose predictor $\mathcal{I}_{\theta_i}$ proposes a target pose for the motion planner to reach before handing control off to the skill policy. It plays a critical role in the HSP framework since it directly dictates the initial state distribution of the skill policy. Even small errors can lead to out-of-distribution states and cause severe performance degradation. To illustrate this, we performed an experiment where we artificially add noise to the initiation pose with increasing scales.
Fig.~\ref{fig:error-succ} plots the success rate of the following skill policy as the noise increases.
As expected, the success rate drops sharply as the prediction error increases. 
In practice, such errors can often occur, especially when the agent has imperfect perception. To mitigate this, we propose two methods that utilize online interactions during deployment and in training.

\textbf{Real-time replanning.} As the robot arm approaches the next skill, more information can be gathered to refine the initiation pose. Typically, a robot equipped with a wrist camera can only capture the target location when the end-effector is close to it. Therefore, instead of running the pose predictor once, we run it at every timestep of the transit phase, while executing a motion plan. 
When the difference between the newly predicted pose and the previously planned target $T_W^{E'_0}$ exceeds a given threshold, we then restart motion planning with the updated pose.

\textbf{\rebuttal{Distillation from a privileged predictor}{Student predictor from a privileged teacher}.} The replanning approach stresses the importance of continuously making predictions during motion planning. However, when using a suboptimal predictor, the motion planning trajectory will inevitably encounter out-of-distribution states, thereby hindering prediction accuracy. During training, we have access to privileged object state information that allows us to use an accurate teacher predictor. Specifically, the privileged predictor selects one of the source demos and transforms the target pose in the demo according to the current object state, similar to the data generation procedure described in Sec.~\ref{sec:data-gen}. We denote the teacher predictor $\mathcal{I}^\text{Priv}$, and the HSP variant that uses this predictor as \emph{HSP-Priv} (HSP-Class in \cite{garrett2024skillmimicgen}).
Given the teacher predictor $\mathcal{I}^\text{Priv}$, we can distill $\mathcal{I}_{\theta_i}$ by minimizing the mean squared error to the teacher predicted pose on the motion planning trajectory states $s\in\tau_\text{MP}$.
We initialize $\mathcal{I}_{\theta_i}$ by sampling trajectories $\tau_\text{MP}$ using $\mathcal{I}^\text{Priv}$ as pose predictor. With a reasonably accurate $\mathcal{I}_{\theta_i}$, we then perform online distillation by sampling $\tau_\text{MP}$ directly with $\mathcal{I}_{\theta_i}$ to bridge the distribution gap.

\subsection{Skill Policies}
\label{sec:alg:policy}

A skill policy $\pi_i$ controls the system for the duration of the skill
until the termination condition is satisfied. We train $\pi_i$ using episodic RL.
The RL reset distribution is the distribution of final states following motion planner execution to reach an initiation pose (Sec.~\ref{sec:alg:init-cond}).
The 0-1 reward function $r_i(s):=[s\in\mathcal{T}_i]$ is determined by the ground-truth termination condition $\mathcal{T}_i$ (Sec.~\ref{sec:alg:term-cond}). 

Although skill learning can be modeled as a standard RL task,
the sparse termination condition reward presents
exploration challenges, which can even lead to complete learning failure~\citep{Stadie2015IncentivizingEI, Tang2017ExplorationAS, Ecoffet2021FirstRT,zhou2024spire}. In light of this, we adopt a residual RL regime~\citep{Johannink2018ResidualRL, Silver2018ResidualPL,zhou2024spire}, starting from a base policy and then fine-tuning it by training an RL agent that outputs differences to the base policy actions, named the residual policy. Specifically, let $\pi^\text{base}$ be the base policy and $\pi^{\text{res}}_{\theta_i}$ be the residual policy to learn. We structure our skill policy as:
$\pi_{\theta_i}(o_t)=\pi^\text{base}_i(o_t)+\pi^{\text{res}}_{\theta_i}(o_t)$.
We then train the residual policy with off-the-shelf reinforcement learning algorithms to maximize the regularized objective:
$\mathcal{J}(\theta_i) = \Esubarg{\tau_{\text{skill}} \sim \pi_{\theta_i}}{T_i(s_t) - \alpha\cdot \|\pi^\text{Res}_{\theta_i}(o_t)\|^2}$.
Here, $\alpha$ is a coefficient controlling the regularization strength, which constrains deviations between the base and the fine-tuned policy~\citep{zhou2024spire} by penalizing the squared L2 norm of the residual actions; $T_i$ is the termination condition for skill fine-tuning (c.f. Sec.~\ref{sec:alg:term-cond}).
We choose the skill policy in HSP (Sec.~\ref{sec:hsp}) as our base policy, which is a behavior-cloning agent trained from the dataset obtained with object-centric data generation. 

\subsection{Termination Classification}
\label{sec:alg:term-cond}

A skill termination condition $\mathcal{T}_i$ determines whether the goal of the current stage has been achieved (i.e., current state $s\in\mathcal{T}_i$) so that the agent can either enter the next stage or terminate.
For deployment, we train a parameterized termination predictor $\mathcal{T}_{\theta_i}:\mathcal{O}\rightarrow\{0, 1\}$ from \AlgoName{} rollouts with a binary cross entropy loss using ground-truth labels $[s_t\in\mathcal{T}_i]$ along skill trajectory $\tau_\text{skill}$.

Compared with the initiation pose predictor in Sec.~\ref{sec:alg:init-cond}, an important distinction is that we do not assume limited observability for the termination predictor during training. 
We opt to use a hand-crafted termination for skill RL training. This is because RL constantly updates the skill policy, which changes the state distribution for the termination predictor. This leads to out-of-distribution predictions that can be exploited by RL.

\textbf{Termination fine-tuning.} 
A termination predictor can make two types of errors: false positives, where a termination is predicted but the subtask is not completed, and false negatives, where a subtask is completed but the predictor fails to declare so. False negatives result in a stricter training condition for the RL agent, but empirically have a limited impact on performance because the policy can still explore other states until the predictor produces a true positive.
However, false positives can lead to \emph{irrecoverable} states~(App. Fig.~\ref{fig:fine-tune}c) and can be exploited by RL training. 

Although we could formulate termination prediction itself as an RL problem with a binary action space, we take the following simpler strategy to reduce false positives.
Let $T_i:\mathcal{S}\rightarrow\{0, 1\}$ be the termination conditions used for skill fine-tuning.
Let $\theta^*_i$ be the parameters fine-tuned with $T_i$. We train a predictor $p_i:\mathcal{S}\rightarrow[0,1]$ to estimate the probability of task success should we terminate the stage at the current state. Formally, $p_i$ is trained to fit $P(s_T\in\mathcal{T} \mid T_i(s_t)=1,\tau\sim\pi_{\theta_i})$, $\mathcal{T}$ being the task completion condition.
Under the RL formulation of stage terminations, $p_i$ represents the Q-function with $T_i$ as the policy.
Let $\mathcal{T}^r_i:\mathcal{S}\rightarrow\{0, 1\}$ be the hand-crafted termination conditions and let $T_i$ initialized to $\mathcal{T}^r_i$. 
Instead of doing RL exploration for an optimal $T_i$, we use a greedy solution by rejecting terminations with a Q-function (i.e., $p_i$) too low.
The new termination conditions with a rejection threshold $\epsilon_{\text{term}}$ is:
$T_i(s):= \mathcal{T}^r_i(s)\cdot\left[p_i(s)>\epsilon_{\text{term}}\right]$.

\subsection{\AlgoName{} Hybrid Policy} 
\label{sec:alg:deployment}

After learning skill initiation conditions, policies, and termination conditions, we can deploy these skills in sequence, using motion planning to connect adjacent pairs of skills.
\rebuttal{
Algorithm~\ref{alg:policy} in Appendix~\ref{app:reprod} displays the \AlgoName{} system pseudocode at deployment time. It iterates through each of the $n$ parameterized skills $\psi_{\theta_i}$.
For the $i$-th skill, it predicts an end-effector skill initialization pose $p$ from the most recent observation $o$ .
Using the motion planner, it plans a trajectory $\tau$ to reach this pose and executes it, possibly while holding an object.
Next, it controls the system using the skill policy $\pi_{\theta_i}$ by predicting and executing actions $a$ until it reaches an observation $o$ that is predicted to correspond to a termination state.
}{
\AlgoName{} iterates through each of the $n$ parameterized skills $\psi_{\theta_i}$.
For the $i$-th skill, it predicts an end-effector skill initialization pose $p$ from the most recent observation $o$ (Sec.~\ref{sec:alg:init-cond}).
Using the motion planner, it plans a trajectory $\tau$ to reach this pose and executes it, while updating the pose target prediction during execution. When the difference between the updated pose and the current motion planner pose exceeds a threshold, \AlgoName{} triggers the motion planner to replan to the new pose (Sec.~\ref{sec:alg:init-cond}).
Next, it controls the system using the fine-tuned skill policy $\pi_{\theta_i}$ (Sec.~\ref{sec:alg:policy}) by predicting and executing actions $a$ until it reaches an observation $o$ that is predicted to correspond to a termination state (Sec.~\ref{sec:alg:term-cond}).
We present the pseudocode for this process in App.~\ref{app:reprod}, Algo.~\ref{alg:policy}.
}

\subsection{End-to-End Distillation} 
\label{sec:alg:distill}

The hybrid policy in Sec.~\ref{sec:alg:deployment} leverages a motion planner to stitch together skill segments.
Motion planning requires a model of the world's collision volume, which can be object meshes if the world is fully observable or a point cloud if it is not~\citep{sundaralingam2023curobo}.
In some deployments, we may wish to bypass the motion planner due to these additional requirements and learn a single end-to-end visuomotor policy.
In \AlgoName{}, we have the flexibility to do either, where there is a tradeoff between additional requirements and learning difficulty, which a user can tailor to their application.

To train an end-to-end policy, we first configure the motion planner to use the same controller as the skill policy, i.e., by switching from joint-space to task-space control.
Then, we generate consistent end-to-end trajectories by stitching together the motion planning and skill policy segments. 

Compared to using baseline imitation agents as the generator, incorporating online data allows \AlgoName{} to produce higher-quality source trajectories. Moreover, the random exploration and exploitation process in RL-based fine-tuning makes \AlgoName{} agents naturally more resistant to deviations from optimal behaviors. This is especially important in the end-to-end distillation setting, since the deviations compound throughout the long-horizon execution.

\section{Experiments}

\textbf{Tasks.} We benchmark \AlgoName{} on several robosuite~\citep{robosuite2020} manipulation tasks, specifically the D2 variants~\citep{garrett2024skillmimicgen} (largest initialization range) of the \emph{Nut Assembly}, \emph{Threading}, \emph{Three Piece}, \emph{Coffee}, and \emph{Coffee Preparation} tasks.
Additionally, we experiment on two real-robot tasks to investigate \AlgoName{}'s capacity for zero-shot real-world deployment.  See Appendix~\ref{app:task} for more detail.

\textbf{Observation space.} The observation space comprises $84\times 84$ RGB images from a front-view camera and a robot wrist camera as well as robot proprioception, namely its end-effector pose and gripper joint position.
For the \emph{Threading} task, we include an additional side-view camera.

\textbf{Demonstrations.} We collected 10 human source demonstrations per task.
Using \AlgoName{}, we automatically adapt these source demonstrations into a BC dataset of 1,000 demonstrations.

\textbf{Baselines.} We use HSP-Priv (Sec.~\ref{sec:alg:init-cond}, \cite{garrett2024skillmimicgen}) as our baseline with privileged information. We further fine-tune its skill policies to establish an upper-bound on performance (ReinforceGen-Priv). 
For baselines sharing the same assumptions, we use HSP.

\textbf{Details.} We skip fine-tuning on stages that have high baseline success rates to save computational cost. For an exhaustive implementation list, refer to App.~\ref{app:ft-impl}. We use hand-crafted stage terminations in our evaluations for all hybrid policies despite it requiring state information, which is in line with prior works~\citep{garrett2024skillmimicgen, zhou2024spire}. We instead include an ablation that uses learned terminations.

\subsection{Simulated Results}
\label{sec:exp:main}

\begin{table}[t]
    \centering
        \begin{footnotesize} 
        \begin{tabular}{|l|rrrrr|r|}
        \toprule
        \textbf{Success Rate (\%)} & \textbf{Nut Assem.} & \textbf{Threading} & \textbf{Three Piece} & \textbf{Coffee} & \textbf{Coffee Prep.} & \textbf{Overall} \\
        \midrule
        \spire{} (Zhou et al.) & N/A & N/A & 86.00 & 98.00 & 84.00 & N/A \\ 
        \midrule
        HSP-Priv (Garrett et al.) & 86.20 & 54.27 & 70.77 & 88.12 & 65.60 & 72.99 \\
        ReinforceGen-Priv (Ours) & \textbf{87.20} & \textbf{89.40} & \textbf{82.24} & \textbf{97.03} & \textbf{77.80} & \textbf{86.66} \\
        \midrule
        HSP (Garrett et al.) & 40.52 & 50.20 & 41.52 & 55.38 & 35.80 & 44.68 \\
        HSP + Replan & 78.40 & 49.80 & 65.80 & 80.20 & 60.20 & 66.88 \\
        \AlgoName{} (Ours) & \textbf{85.80} & \textbf{82.20} & \textbf{80.40} & \textbf{93.81} & \textbf{80.80} & \textbf{84.60} \\
        \bottomrule
    \end{tabular}
    \end{footnotesize}
    \caption{Comparison of task success rates across all tasks. \spire{} uses 200 demonstrations while the rest use 10. SPIRE, HSP-Priv (Sec.~\ref{sec:alg:init-cond}), and ReinforceGen-Priv use privileged state information, while the rest rely only on observations. The numbers are averaged from over 500 rollouts except for \spire{}, which uses 50.}
    \label{tab:main-result}
\end{table}

\textbf{\AlgoName{} is proficient in all tasks despite partial observability.} Tab.~\ref{tab:main-result} shows that \AlgoName{} achieves over \textbf{80\%} task completion rate in all our tested tasks, including long-horizon tasks with as many as 5 stages, all while relying only on camera sensory input and proprioception information. 

\textbf{\AlgoName{} achieves an 89\% relative performance increase over baselines without state observability on average.} In Tab.~\ref{tab:main-result}, comparing HSP-Priv and HSP, we observe a significant drop in success rates due to the lack of object location awareness. With the same accessible information, \AlgoName{} nearly doubles HSP's overall success rate. We also notice that our gains are especially large in longer-horizon tasks with 4-5 stages, i.e., \emph{Nut Assembly}, \emph{Three Piece}, \emph{Coffee Preparation}, reaching \textbf{109\%} relative increase.

\textbf{\AlgoName{} is competitive with baselines with privileged state information.} We fine-tuned HSP-Priv with access to state information as a pseudo-performance upper bound (\AlgoName{}-Priv in Tab.~\ref{tab:main-result}). Despite the disadvantages, \AlgoName{} only shows a maximum deficit of \textbf{8\%} in \emph{Threading}, with \textbf{2\%} decrease in the overall success rate. that improves hierarchical data generation by incorporating online exploration and environmental feedback using RL.

\begin{figure}[t]
\vspace{-6mm}
     \centering
     \begin{subfigure}[b]{0.24\textwidth}
         \centering
         \includegraphics[width=.82\textwidth]{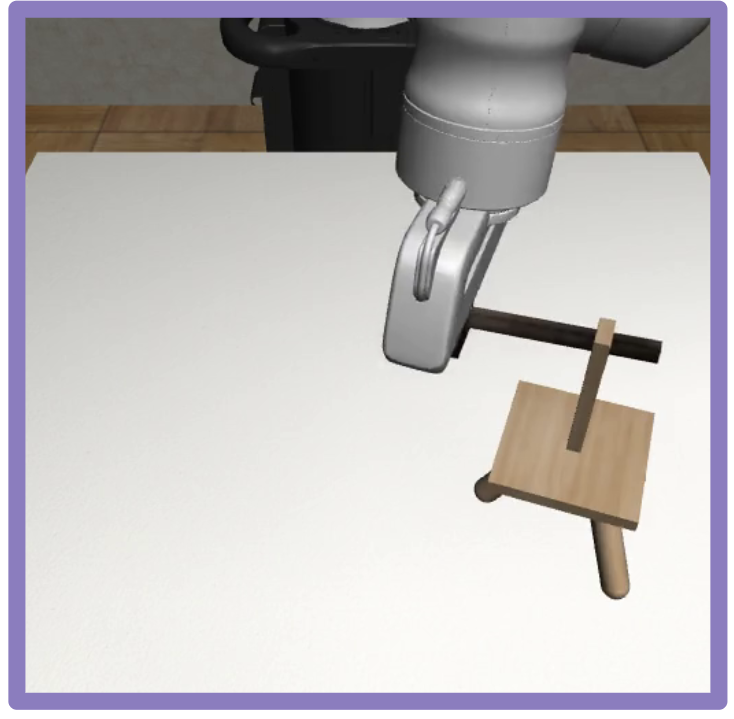}
         \caption{\centering\emph{Threading} (Stage 2) - \textbf{82.2\%}}
         \label{fig:qualitative-threading}
     \end{subfigure}
     \hfill
     \begin{subfigure}[b]{0.24\textwidth}
         \centering
         \includegraphics[width=.82\textwidth]{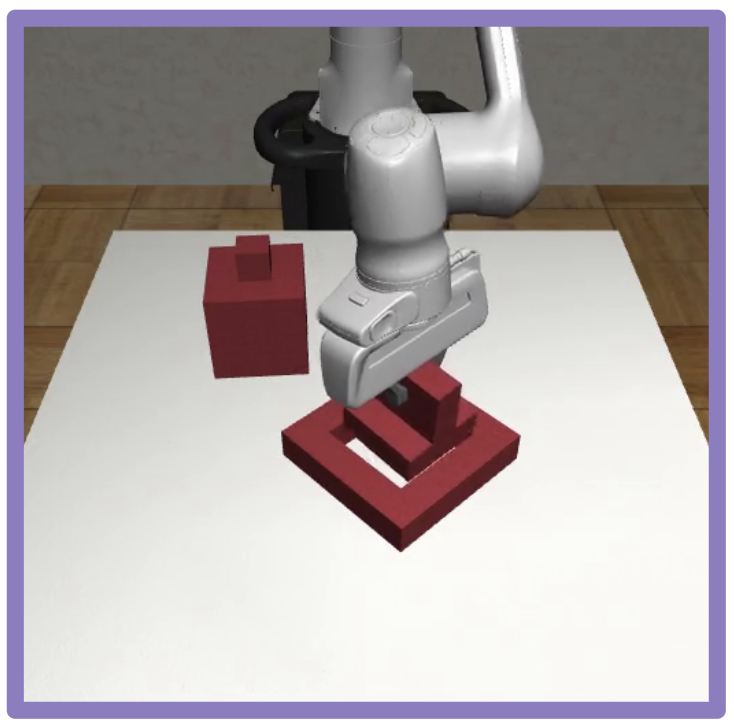}
         \caption{\centering\emph{Three Piece} (Stage 2) - \textbf{93.8\%}}
         \label{fig:qualitative-threepiece}
     \end{subfigure}
     \hfill
     \begin{subfigure}[b]{0.24\textwidth}
         \centering
         \includegraphics[width=.82\textwidth]{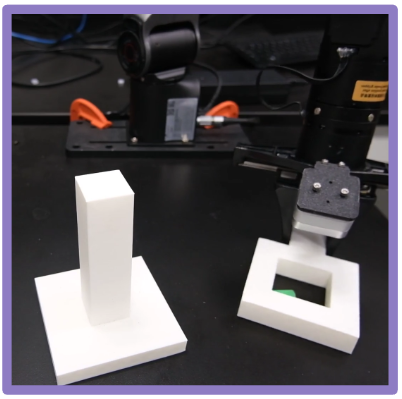}
         \caption{\centering\emph{Square} (Real) -\\ \textbf{70\%}}
         \label{fig:qualitative-square}
     \end{subfigure}
     \begin{subfigure}[b]{0.24\textwidth}
         \centering
         \includegraphics[width=.82\textwidth]{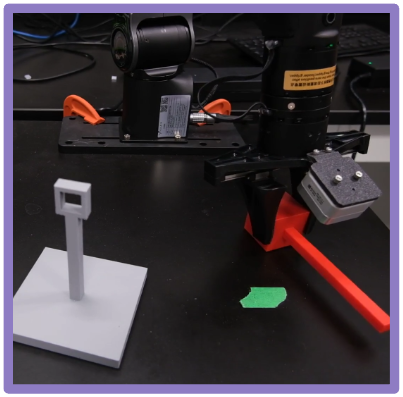}
         \caption{\centering\emph{Threading} (Real) -\\ \textbf{35\%}}
         \label{fig:qualitative-threading-real}
     \end{subfigure}
     \vspace{-1mm}
        \caption{\AlgoName{} agents complete high-precision skills with high success rates (numbers shown in bold).} 
        \label{fig:qualitative}
\end{figure}

\textbf{Pose target replanning improves motion planning reaching accuracy and skill completion.} Comparing HSP and HSP+Replan in Table~\ref{tab:main-result}, adding replanning alone increases the overall success rate of HSP by \textbf{22\%} (from 44.68\% to 66.88\%), a relative \textbf{50\%} improvement. 
We present a more detailed case study in \emph{Nut Assembly} in Appendix~\ref{app:ablation-replan}.

\begin{table}[h]
\centering
\vspace{-3mm}
\begin{footnotesize}
\begin{tabular}{|l|rrrr|}
      \toprule
\textbf{Improvement (\%)} & \emph{Nut Assembly} & \emph{Threading} & \emph{Three Piece} & \emph{Coffee} \\
      \midrule
Success Rate & 4.62 & 65.73 & 10.55 & 16.74\\
Efficiency  & 8.43 & 16.39 & 10.18 & 3.97 \\
    \bottomrule
  \end{tabular}
    \caption{
    \AlgoName{}'s policy improvement through skill fine-tuning. 
    Efficiency is the average number of timesteps to complete each skill.
    Success Rate and Efficiency are averaged over all task stages.
    }
\vspace{-5mm}
    \label{tab:skill-ft-overall}
\end{footnotesize}
\end{table}
\textbf{Skill policy fine-tuning recovers inaccurate starting poses and suboptimal imitated skills.} We ablate skill fine-tuning from \AlgoName{} to show its effectiveness in Tab.~\ref{tab:skill-ft-overall}. On average, skill fine-tuning improves task completion rates by \textbf{24.41\%} while using \textbf{9.74\%} fewer steps to complete. Skill fine-tuning is most effective in the second stage of \emph{Threading}, which is the most precision-demanding stage among all benchmarked tasks, and where the IL agent has the lowest success rate.

\textbf{Termination fine-tuning repairs cross-stage causal effects.} When directly applied to HSP-Priv, termination fine-tuning improves its success rate from 66.44\% to 70.77\%; after fine-tuning skills with the new terminations, the success rate further increases from 72.60\% to 82.24\%.

\begin{table}[h]
\vspace{-4mm}
\centering
\begin{footnotesize}
\begin{tabular}{|l|rrrr|}
      \toprule
\textbf{Success Rate (\%)} & \emph{Nut Assembly} & \emph{Threading} & \emph{Three Piece} & \emph{Coffee} \\
      \midrule
HSP-Priv & \textbf{35.00} & 60.40 & \textbf{20.20} & 84.60 \\
\AlgoName{}-Priv & 28.00 & \textbf{83.60} & 18.60 & \textbf{94.20} \\
    \bottomrule
  \end{tabular}
    \caption{
    End-to-end distillation results using datasets generated with \AlgoName{}-Priv and HSP-Priv.
    }
    \label{tab:distillation}
\vspace{-3mm}
\end{footnotesize}
\end{table}

\textbf{\AlgoName{} can train end-to-end policies.} We also distill the hybrid \AlgoName{} agents to end-to-end visuomotor policies. We first configure the motion planner's action space to be consistent with the skill policy, allowing us to stitch the connect and skill phase trajectories of each stage together. We then create a dataset of such rollouts and train a BC agent from them. The resulting success rates are shown in Table~\ref{tab:distillation}.
In \emph{Threading} and \emph{Coffee}, \AlgoName{} distills to proficient end-to-end agents, with significant improvement over the baseline. However, in \emph{Nut Assembly} and \emph{Three Piece}, both agents struggle to complete the task. We hypothesize that these tasks require long transit motions, which exacerbate the difficulty of learning when compared to learning just contact-rich motion.

\subsection{Real World Evaluation}

We demonstrate the real-world deployment capability of \AlgoName{} on a variant of the \textit{Square} task (Fig.~\ref{fig:qualitative-square}) using a 6-DOF AgileX PiPER arm and a RealSense D435i camera.
We collect 10 source human teleoperation demonstrations with a real robot and replay the trajectories in simulation, using FoundationPose~\citep{foundationposewen2024} to estimate the initial object state.
Because we already assume pose estimation for real-to-sim demonstration transfer, we skipped initialization pose prediction (Sec.~\ref{sec:alg:init-cond}) and instead simply adapted ground-truth demonstration initialization poses.
Agents are evaluated for 20 rollouts. An additional evaluation on the real-world \textit{Threading} task is presented in App.~\ref{sec:app:real-results-threading}.

\textbf{\AlgoName{} successfully transfers to real robots.} From Table~\ref{tab:sim2real}, \AlgoName{} agents reach a \textbf{70\%} success rate in the real world, a \textbf{2.8$\times$} relative performance improvement compared to our BC baseline (HSP), which is consistent with the results of the more expansive simulation experiments.

\textbf{Termination fine-tuning reduces reward exploitation in skill fine-tuning.} We ablate \AlgoName{} with respect to termination fine-tuning~(Sec.~\ref{sec:alg:term-cond}). In simulation, \AlgoName{} completes grasping stages with almost perfect success rates at the cost of drastic decreases in second stage success rates. 
Tab.~\ref{tab:sim2real} shows this also occurs in the real world, as we see lower sim-to-real transfer compared to HSP, and eventually a regression on real-world success rate. 
We further evaluate a version with hand-crafted conditions for stable grasps (Term. w/ Human FT) that achieves almost perfect sim-to-real transfer, which demonstrates the importance of termination fine-tuning.

\begin{table}[h]
\vspace{-3mm}
\begin{footnotesize}
\centering
\begin{tabular}{|l|rrr|}
      \toprule
\textbf{Success Rate (\%)} & \emph{Simulation Success.} & \emph{Real Success.} & \emph{Sim-to-Real Success.} \\
      \midrule
HSP & 40.00 & 25.00 & 62.50 \\
ReinforceGen (w/o Term. FT) & 45.00 & 15.00 & 33.33 \\
ReinforceGen & \textbf{90.00} & 70.00 & 77.78 \\
ReinforceGen (w/ Human Term. FT) & \textbf{90.00} & \textbf{85.00} & \textbf{94.44} \\
      \bottomrule
\end{tabular}
\caption{Real-world evaluation results. Success rates are averaged over 20 rollouts. Sim-to-real transfer success rates are calculated by dividing real-world success rate by simulation success rate.
}
\label{tab:sim2real}
\end{footnotesize}
\vspace{-5mm}
\end{table}
\section{Conclusion and Limitations}

We presented \AlgoName{}, an automated demonstration generation system that integrates planning, demonstration adaptation, and reinforcement learning to bootstrap a small set of human demonstrations into a large high-quality dataset.
These demonstrations can be used to train end-to-end imitation learners or atomic skill policies that are stitched together using motion planning, demonstrating the flexibility of our approach.
The use of reinforcement learning for exploration improves policy success rates when compared to prior works due to its ability to explore beyond adapted demonstrations.

\textbf{Limitations.} Our demonstration system is not fully automated as we still assume access to a handful of human demonstrations (e.g. 1 - 10).
In order to adapt skill demonstrations, we assume access to reference frame labeled per skill, which might be difficult to specify for skills that don't involve conventional objects, for example a sweep skill involving granual media.
In our experiments, we train specialized task-conditioned policies.
Future work involves investigating the extent to which skill policies can be used across tasks.
We use the observed point cloud as the collision volume for motion planning, which might be the insufficient in situations with heavy partial observability.

\bibliography{main}

\begin{thebibliography}{33}
\providecommand{\natexlab}[1]{#1}
\providecommand{\url}[1]{\texttt{#1}}
\expandafter\ifx\csname urlstyle\endcsname\relax
  \providecommand{\doi}[1]{doi: #1}\else
  \providecommand{\doi}{doi: \begingroup \urlstyle{rm}\Url}\fi

\bibitem[Mandlekar et~al.(2021)Mandlekar, Xu, Wong, Nasiriany, Wang, Kulkarni,
  Fei-Fei, Savarese, Zhu, and Mart\'{i}n-Mart\'{i}n]{mandlekar2021matters}
A.~Mandlekar, D.~Xu, J.~Wong, S.~Nasiriany, C.~Wang, R.~Kulkarni, L.~Fei-Fei,
  S.~Savarese, Y.~Zhu, and R.~Mart\'{i}n-Mart\'{i}n.
\newblock What matters in learning from offline human demonstrations for robot
  manipulation.
\newblock In \emph{Conference on Robot Learning (CoRL)}, 2021.

\bibitem[Mandlekar et~al.(2023)Mandlekar, Nasiriany, Wen, Akinola, Narang, Fan,
  Zhu, and Fox]{mimicgen}
A.~Mandlekar, S.~Nasiriany, B.~Wen, I.~Akinola, Y.~Narang, L.~Fan, Y.~Zhu, and
  D.~Fox.
\newblock Mimicgen: A data generation system for scalable robot learning using
  human demonstrations.
\newblock In \emph{Conference on Robot Learning (CoRL)}, 2023.

\bibitem[Garrett et~al.(2024)Garrett, Mandlekar, Wen, and
  Fox]{garrett2024skillmimicgen}
C.~R. Garrett, A.~Mandlekar, B.~Wen, and D.~Fox.
\newblock Skillmimicgen: Automated demonstration generation for efficient skill
  learning and deployment.
\newblock In \emph{8th Annual Conference on Robot Learning}, 2024.
\newblock URL \url{https://openreview.net/forum?id=YOFrRTDC6d}.

\bibitem[Jiang et~al.(2025)Jiang, Xie, Lin, Xu, Wan, Mandlekar, Fan, and
  Zhu]{jiang2024dexmimicen}
Z.~Jiang, Y.~Xie, K.~Lin, Z.~Xu, W.~Wan, A.~Mandlekar, L.~Fan, and Y.~Zhu.
\newblock Dexmimicgen: Automated data generation for bimanual dexterous
  manipulation via imitation learning.
\newblock In \emph{2025 IEEE International Conference on Robotics and
  Automation (ICRA)}, 2025.

\bibitem[McDonald and Hadfield-Menell(2022)]{mcdonald2022guided}
M.~J. McDonald and D.~Hadfield-Menell.
\newblock Guided imitation of task and motion planning.
\newblock In \emph{Conference on Robot Learning}, pages 630--640. PMLR, 2022.

\bibitem[Dalal et~al.(2023)Dalal, Mandlekar, Garrett, Handa, Salakhutdinov, and
  Fox]{dalal2023imitating}
M.~Dalal, A.~Mandlekar, C.~Garrett, A.~Handa, R.~Salakhutdinov, and D.~Fox.
\newblock Imitating task and motion planning with visuomotor transformers.
\newblock \emph{arXiv preprint arXiv:2305.16309}, 2023.

\bibitem[Garrett et~al.(2021)Garrett, Chitnis, Holladay, Kim, Silver,
  Kaelbling, and Lozano-P{\'e}rez]{garrett2021integrated}
C.~R. Garrett, R.~Chitnis, R.~Holladay, B.~Kim, T.~Silver, L.~P. Kaelbling, and
  T.~Lozano-P{\'e}rez.
\newblock Integrated task and motion planning.
\newblock \emph{Annual review of control, robotics, and autonomous systems},
  4:\penalty0 265--293, 2021.

\bibitem[Mandlekar et~al.(2023)Mandlekar, Garrett, Xu, and
  Fox]{mandlekar2023hitltamp}
A.~Mandlekar, C.~Garrett, D.~Xu, and D.~Fox.
\newblock Human-in-the-loop task and motion planning for imitation learning.
\newblock In \emph{7th Annual Conference on Robot Learning}, 2023.

\bibitem[Zhou et~al.(2024)Zhou, Garg, Fox, Garrett, and
  Mandlekar]{zhou2024spire}
Z.~Zhou, A.~Garg, D.~Fox, C.~R. Garrett, and A.~Mandlekar.
\newblock {SPIRE}: Synergistic planning, imitation, and reinforcement learning
  for long-horizon manipulation.
\newblock In \emph{8th Annual Conference on Robot Learning}, 2024.
\newblock URL \url{https://openreview.net/forum?id=cvUXoou8iz}.

\bibitem[Florence et~al.(2022)Florence, Lynch, Zeng, Ramirez, Wahid, Downs,
  Wong, Lee, Mordatch, and Tompson]{florence2022implicit}
P.~Florence, C.~Lynch, A.~Zeng, O.~A. Ramirez, A.~Wahid, L.~Downs, A.~Wong,
  J.~Lee, I.~Mordatch, and J.~Tompson.
\newblock Implicit behavioral cloning.
\newblock In \emph{Conference on Robot Learning}, pages 158--168. PMLR, 2022.

\bibitem[Chi et~al.(2023)Chi, Feng, Du, Xu, Cousineau, Burchfiel, and
  Song]{diffusionpolicy}
C.~Chi, S.~Feng, Y.~Du, Z.~Xu, E.~Cousineau, B.~Burchfiel, and S.~Song.
\newblock Diffusion policy: Visuomotor policy learning via action diffusion.
\newblock \emph{arXiv preprint arXiv:2303.04137}, 2023.

\bibitem[Shafiullah et~al.(2022)Shafiullah, Cui, Altanzaya, and Pinto]{bet}
N.~M. Shafiullah, Z.~Cui, A.~A. Altanzaya, and L.~Pinto.
\newblock Behavior transformers: Cloning $ k $ modes with one stone.
\newblock \emph{Advances in neural information processing systems},
  35:\penalty0 22955--22968, 2022.

\bibitem[Zhao et~al.(2023)Zhao, Kumar, Levine, and Finn]{act}
T.~Z. Zhao, V.~Kumar, S.~Levine, and C.~Finn.
\newblock Learning fine-grained bimanual manipulation with low-cost hardware,
  2023.

\bibitem[Kim et~al.(2024)Kim, Pertsch, Karamcheti, Xiao, Balakrishna, Nair,
  Rafailov, Foster, Lam, Sanketi, Vuong, Kollar, Burchfiel, Tedrake, Sadigh,
  Levine, Liang, and Finn]{kim24openvla}
M.~Kim, K.~Pertsch, S.~Karamcheti, T.~Xiao, A.~Balakrishna, S.~Nair,
  R.~Rafailov, E.~Foster, G.~Lam, P.~Sanketi, Q.~Vuong, T.~Kollar,
  B.~Burchfiel, R.~Tedrake, D.~Sadigh, S.~Levine, P.~Liang, and C.~Finn.
\newblock Openvla: An open-source vision-language-action model.
\newblock \emph{arXiv preprint arXiv:2406.09246}, 2024.

\bibitem[Tang et~al.(2025)Tang, Abbatematteo, Hu, Chandra, Martín-Martín, and
  Stone]{annurev:/content/journals/10.1146/annurev-control-030323-022510}
C.~Tang, B.~Abbatematteo, J.~Hu, R.~Chandra, R.~Martín-Martín, and P.~Stone.
\newblock Deep reinforcement learning for robotics: A survey of real-world
  successes.
\newblock \emph{Annual Review of Control, Robotics, and Autonomous Systems},
  8\penalty0 (Volume 8, 2025):\penalty0 153--188, 2025.
\newblock ISSN 2573-5144.
\newblock \doi{https://doi.org/10.1146/annurev-control-030323-022510}.
\newblock URL
  \url{https://www.annualreviews.org/content/journals/10.1146/annurev-control-030323-022510}.

\bibitem[Wu et~al.(2022)Wu, Escontrela, Hafner, Goldberg, and
  Abbeel]{Wu2022DayDreamerWM}
P.~Wu, A.~Escontrela, D.~Hafner, K.~Goldberg, and P.~Abbeel.
\newblock Daydreamer: World models for physical robot learning.
\newblock In \emph{Conference on Robot Learning}, 2022.
\newblock URL \url{https://api.semanticscholar.org/CorpusID:250088882}.

\bibitem[Yu et~al.(2023)Yu, Gileadi, Fu, Kirmani, Lee, Gonzalez~Arenas,
  Lewis~Chiang, Erez, Hasenclever, Humplik, Ichter, Xiao, Xu, Zeng, Zhang,
  Heess, Sadigh, Tan, Tassa, and Xia]{yu2023language}
W.~Yu, N.~Gileadi, C.~Fu, S.~Kirmani, K.-H. Lee, M.~Gonzalez~Arenas, H.-T.
  Lewis~Chiang, T.~Erez, L.~Hasenclever, J.~Humplik, B.~Ichter, T.~Xiao, P.~Xu,
  A.~Zeng, T.~Zhang, N.~Heess, D.~Sadigh, J.~Tan, Y.~Tassa, and F.~Xia.
\newblock Language to rewards for robotic skill synthesis.
\newblock \emph{Arxiv preprint arXiv:2306.08647}, 2023.

\bibitem[Nair et~al.(2017)Nair, McGrew, Andrychowicz, Zaremba, and
  Abbeel]{Nair2017OvercomingEI}
A.~Nair, B.~McGrew, M.~Andrychowicz, W.~Zaremba, and P.~Abbeel.
\newblock Overcoming exploration in reinforcement learning with demonstrations.
\newblock \emph{2018 IEEE International Conference on Robotics and Automation
  (ICRA)}, pages 6292--6299, 2017.

\bibitem[Johannink et~al.(2018)Johannink, Bahl, Nair, Luo, Kumar, Loskyll,
  Ojea, Solowjow, and Levine]{Johannink2018ResidualRL}
T.~Johannink, S.~Bahl, A.~Nair, J.~Luo, A.~Kumar, M.~Loskyll, J.~A. Ojea,
  E.~Solowjow, and S.~Levine.
\newblock Residual reinforcement learning for robot control.
\newblock \emph{2019 International Conference on Robotics and Automation
  (ICRA)}, pages 6023--6029, 2018.

\bibitem[Dalal et~al.(2024)Dalal, Chiruvolu, Chaplot, and
  Salakhutdinov]{dalal2024psl}
M.~Dalal, T.~Chiruvolu, D.~Chaplot, and R.~Salakhutdinov.
\newblock Plan-seq-learn: Language model guided rl for solving long horizon
  robotics tasks.
\newblock In \emph{International Conference on Learning Representations}, 2024.

\bibitem[Xue et~al.(2025)Xue, Deng, Chen, Wang, Yuan, and Xu]{xue2025demogen}
Z.~Xue, S.~Deng, Z.~Chen, Y.~Wang, Z.~Yuan, and H.~Xu.
\newblock Demogen: Synthetic demonstration generation for data-efficient
  visuomotor policy learning.
\newblock \emph{arXiv preprint arXiv:2502.16932}, 2025.

\bibitem[Lin et~al.(2025)Lin, Ragunath, McAlinden, Prasad, Wu, Zhu, and
  Bohg]{lin2025constraint}
K.~Lin, V.~Ragunath, A.~McAlinden, A.~Prasad, J.~Wu, Y.~Zhu, and J.~Bohg.
\newblock Constraint-preserving data generation for visuomotor policy learning.
\newblock \emph{arXiv preprint arXiv:2508.03944}, 2025.

\bibitem[Li et~al.(2025)Li, Xu, Bahety, Yin, Jiang, Huang, Wong, Garlanka,
  Gokmen, Zhang, et~al.]{li2025momagen}
C.~Li, M.~Xu, A.~Bahety, H.~Yin, Y.~Jiang, H.~Huang, J.~Wong, S.~Garlanka,
  C.~Gokmen, R.~Zhang, et~al.
\newblock Momagen: Generating demonstrations under soft and hard constraints
  for multi-step bimanual mobile manipulation.
\newblock In \emph{RSS 2025 Workshop on Whole-body Control and Bimanual
  Manipulation: Applications in Humanoids and Beyond}, 2025.

\bibitem[Wang et~al.(2021)Wang, Garrett, Kaelbling, and
  Lozano-P{\'e}rez]{wang2021learning}
Z.~Wang, C.~R. Garrett, L.~P. Kaelbling, and T.~Lozano-P{\'e}rez.
\newblock Learning compositional models of robot skills for task and motion
  planning.
\newblock \emph{The International Journal of Robotics Research}, 40\penalty0
  (6-7):\penalty0 866--894, 2021.

\bibitem[Stolle and Precup(2002)]{Stolle2002LearningOI}
M.~Stolle and D.~Precup.
\newblock Learning options in reinforcement learning.
\newblock In \emph{Symposium on Abstraction, Reformulation and Approximation},
  2002.
\newblock URL \url{https://api.semanticscholar.org/CorpusID:16398811}.

\bibitem[Stadie et~al.(2015)Stadie, Levine, and
  Abbeel]{Stadie2015IncentivizingEI}
B.~C. Stadie, S.~Levine, and P.~Abbeel.
\newblock Incentivizing exploration in reinforcement learning with deep
  predictive models.
\newblock \emph{ArXiv}, abs/1507.00814, 2015.

\bibitem[Tang et~al.(2017)Tang, Houthooft, Foote, Stooke, Chen, Duan, Schulman,
  Turck, and Abbeel]{Tang2017ExplorationAS}
H.~Tang, R.~Houthooft, D.~Foote, A.~Stooke, X.~Chen, Y.~Duan, J.~Schulman,
  F.~D. Turck, and P.~Abbeel.
\newblock \#exploration: A study of count-based exploration for deep
  reinforcement learning.
\newblock \emph{NIPS}, 2017.

\bibitem[Ecoffet et~al.(2021)Ecoffet, Huizinga, Lehman, Stanley, and
  Clune]{Ecoffet2021FirstRT}
A.~Ecoffet, J.~Huizinga, J.~Lehman, K.~O. Stanley, and J.~Clune.
\newblock First return then explore.
\newblock \emph{Nature}, 590 7847:\penalty0 580--586, 2021.

\bibitem[Silver et~al.(2018)Silver, Allen, Tenenbaum, and
  Kaelbling]{Silver2018ResidualPL}
T.~Silver, K.~R. Allen, J.~B. Tenenbaum, and L.~P. Kaelbling.
\newblock Residual policy learning.
\newblock \emph{ArXiv}, abs/1812.06298, 2018.

\bibitem[Sundaralingam et~al.(2023)Sundaralingam, Hari, Fishman, Garrett,
  Van~Wyk, Blukis, Millane, Oleynikova, Handa, Ramos,
  et~al.]{sundaralingam2023curobo}
B.~Sundaralingam, S.~K.~S. Hari, A.~Fishman, C.~Garrett, K.~Van~Wyk, V.~Blukis,
  A.~Millane, H.~Oleynikova, A.~Handa, F.~Ramos, et~al.
\newblock Curobo: Parallelized collision-free robot motion generation.
\newblock In \emph{IEEE International Conference on Robotics and Automation
  (ICRA)}, 2023.
\newblock URL \url{https://arxiv.org/pdf/2310.17274}.

\bibitem[Zhu et~al.(2020)Zhu, Wong, Mandlekar, and
  Mart{\'i}n-Mart{\'i}n]{robosuite2020}
Y.~Zhu, J.~Wong, A.~Mandlekar, and R.~Mart{\'i}n-Mart{\'i}n.
\newblock robosuite: A modular simulation framework and benchmark for robot
  learning.
\newblock In \emph{arXiv preprint arXiv:2009.12293}, 2020.

\bibitem[Bowen et~al.(2024)Bowen, Yang, Kautz, and
  Birchfield]{foundationposewen2024}
W.~Bowen, W.~Yang, J.~Kautz, and S.~Birchfield.
\newblock {FoundationPose}: Unified 6d pose estimation and tracking of novel
  objects.
\newblock In \emph{CVPR}, 2024.

\bibitem[Yarats et~al.(2021)Yarats, Fergus, Lazaric, and
  Pinto]{yarats2021drqv2}
D.~Yarats, R.~Fergus, A.~Lazaric, and L.~Pinto.
\newblock Mastering visual continuous control: Improved data-augmented
  reinforcement learning.
\newblock \emph{arXiv preprint arXiv:2107.09645}, 2021.

\end{thebibliography}

\appendix

\clearpage
\section{Overview}

\begin{itemize}
    \item Appendix~\ref{app:task} shows the details of the benchmark task set.
    \item Appendix~\ref{app:reprod} lists the details to reproduce our results.
    \item Appendix~\ref{app:add-results} includes additional experiment results.
    \item \rebuttal{}{Appendix~\ref{app:details} includes additional implementation details.}
    \item Appendix~\ref{app:real} provides details on our real-world evaluation setup.
\end{itemize}

\clearpage
\section{Tasks}
\label{app:task}

\subsection{Simulation Tasks}

\begin{figure}[ht]
     \centering
     \begin{subfigure}[b]{0.57\textwidth}
         \centering
         \includegraphics[width=\textwidth]{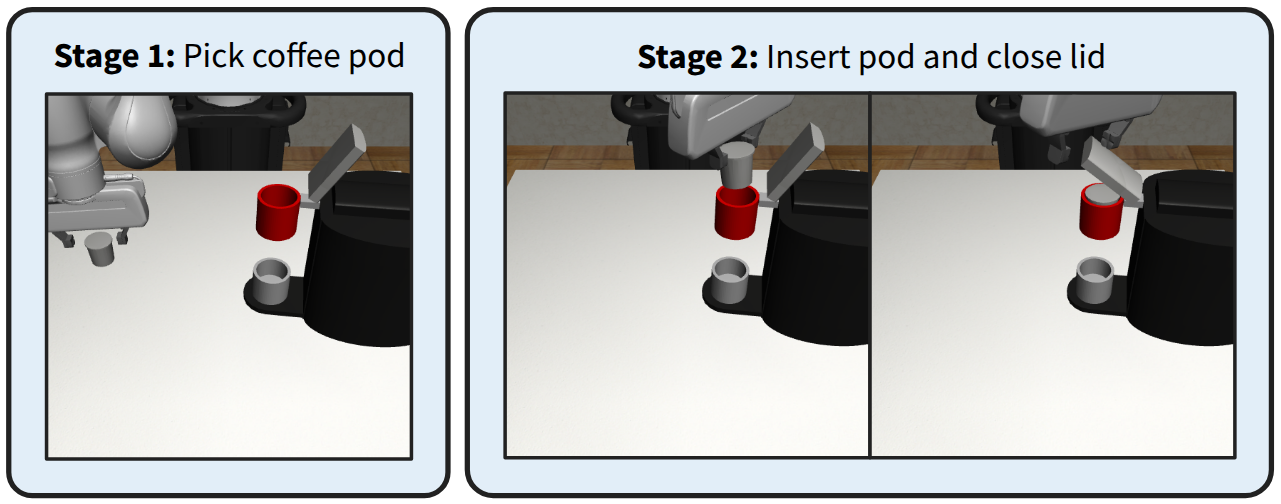}
         \caption{Coffee (2 stages)}
         \label{fig:app:env_coffee}
     \end{subfigure}
     \hfill
     \begin{subfigure}[b]{0.41\textwidth}
         \centering
         \includegraphics[width=\textwidth]{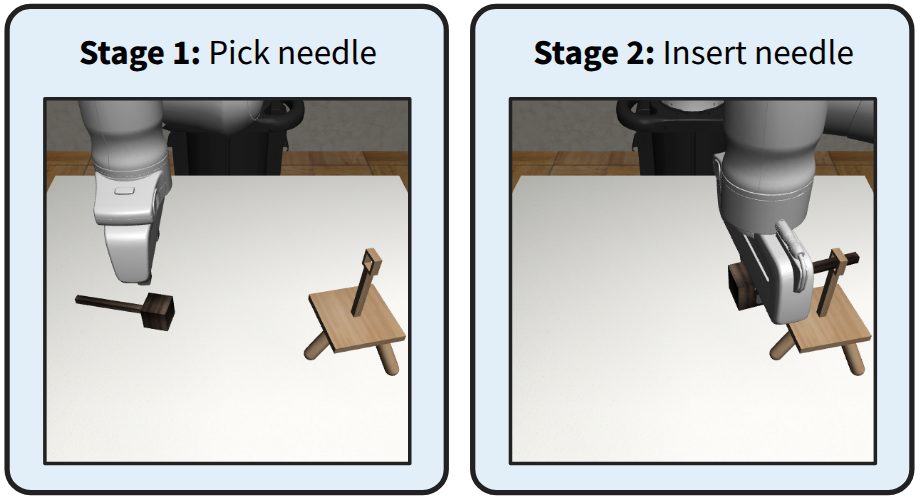}
         \caption{Threading (2 stages)}
         \label{fig:app:env_threading}
     \end{subfigure}
     \hfill
     \begin{subfigure}[b]{0.85\textwidth}
         \centering
         \includegraphics[width=\textwidth]{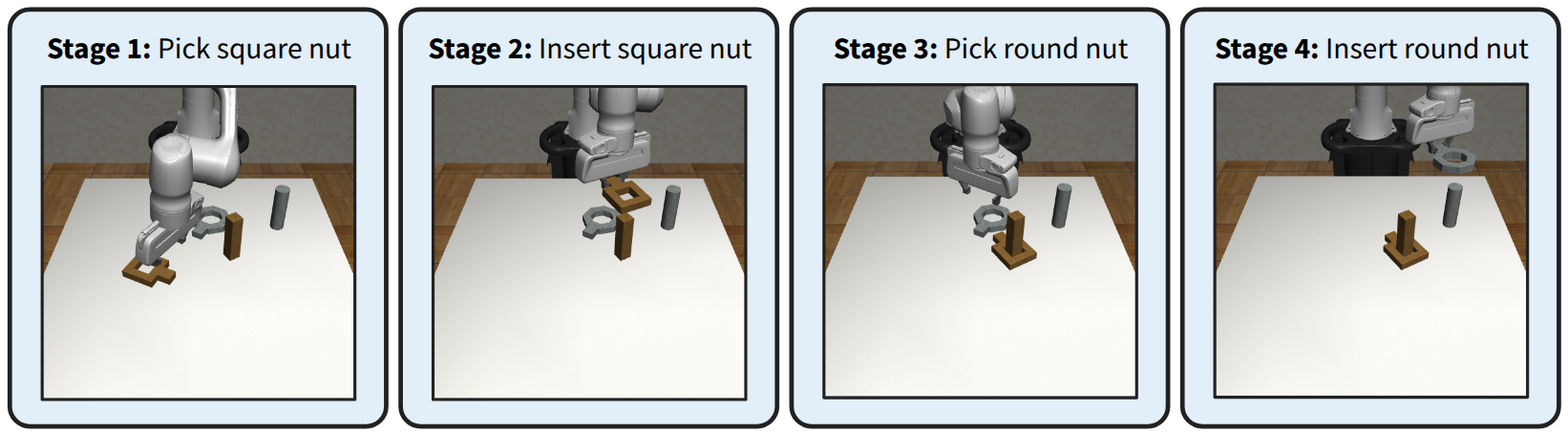}
         \caption{Nut Assembly (4 stages)}
         \label{fig:app:env_nut_assembly}
     \end{subfigure}
     \begin{subfigure}[b]{0.85\textwidth}
         \centering
         \includegraphics[width=\textwidth]{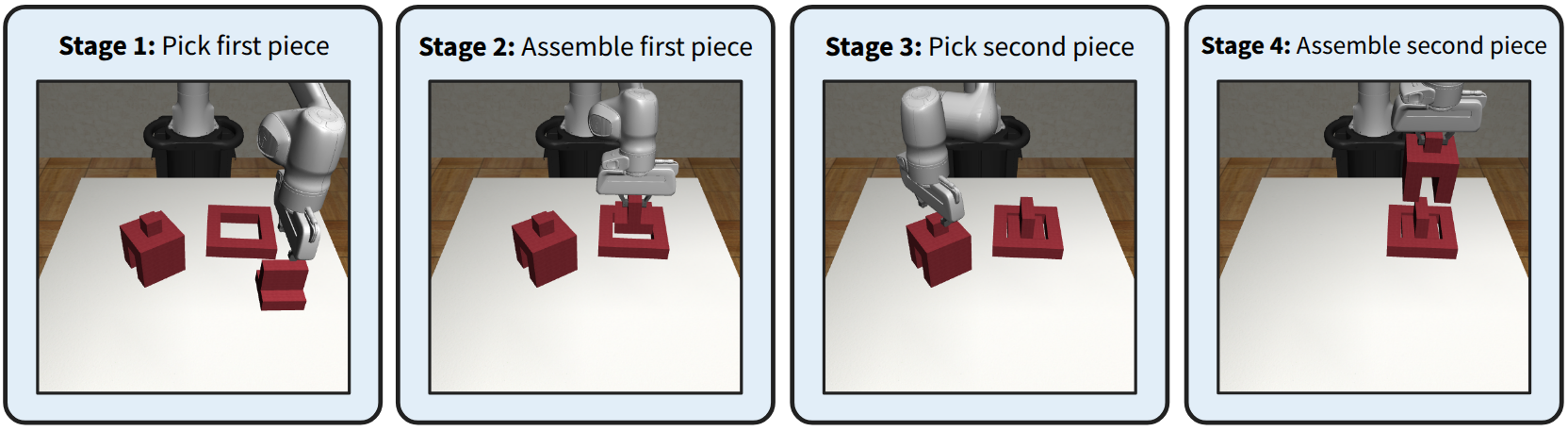}
         \caption{Three Piece Assembly (4 stages)}
         \label{fig:app:env_three_piece}
     \end{subfigure}
     \begin{subfigure}[b]{0.85\textwidth}
         \centering
         \includegraphics[width=\textwidth]{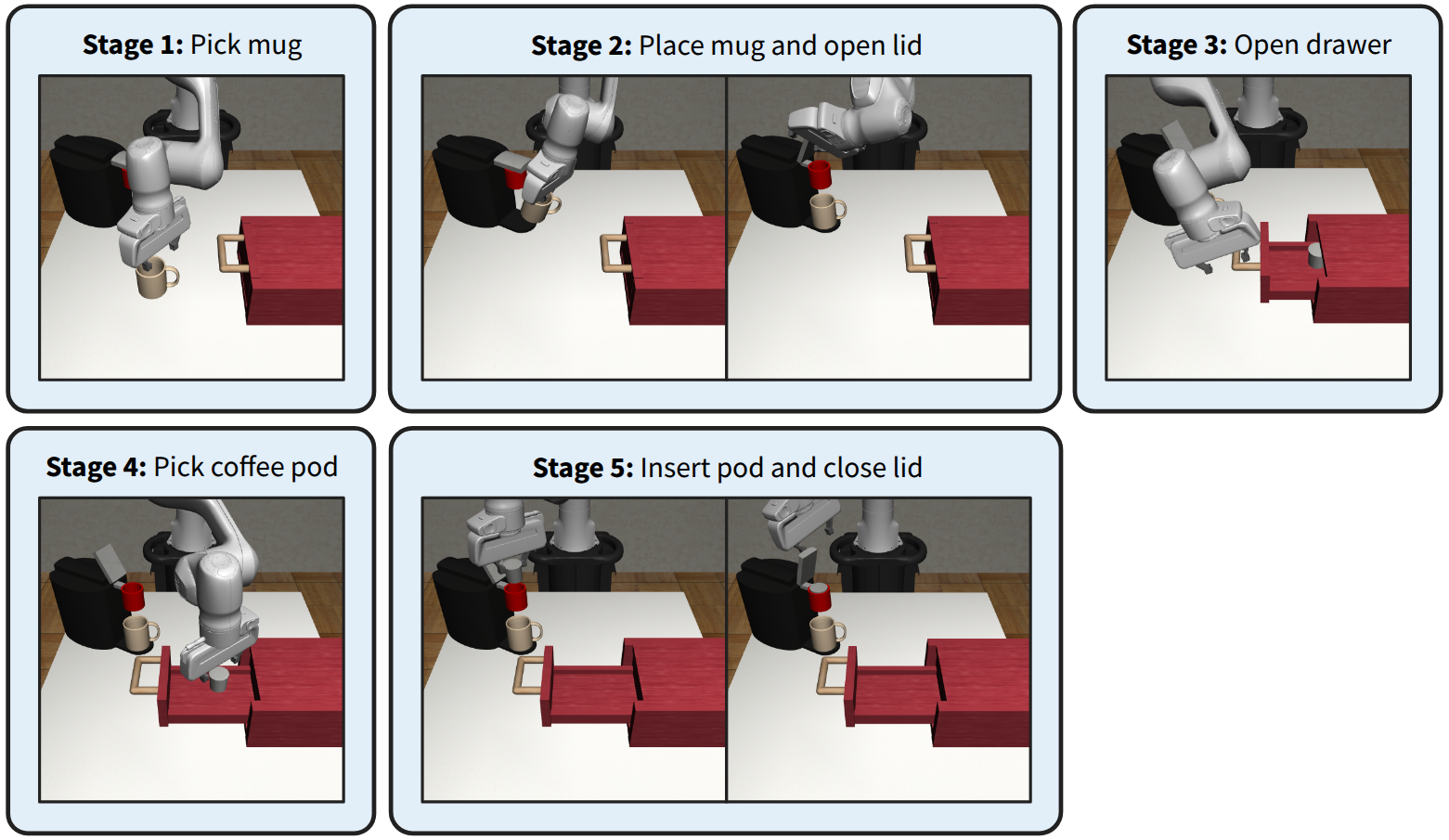}
         \caption{Coffee Preparation (5 stages)}
         \label{fig:app:env_coffee_full}
     \end{subfigure}
        \caption{All five simulation tasks showcased.}
        \label{fig:app:task_showcase}
\end{figure}

We choose the five tasks from Robosuite~\citep{robosuite2020,garrett2024skillmimicgen} (Fig.~\ref{fig:app:task_showcase}). We use the largest initiation range version (D2) for all tasks. The only exception is \emph{Nut Assembly}, where we reduce the $x$-range of the nuts' placement to $(-0.15, 0.15)$ since the original range produces unreachable initial positions.

\textbf{Observations.} The observation consists of two (or three, in Threading) 84$\times$84 RGB images and robot proprioception state, including a 6-dim end-effector pose composed of 3-dim Euclidean position in meters and 3-dim rotation represented in axis-angle form; a 2-dim gripper position, representing the distances of the two grippers to the center; and a 7-dimensional joint configuration. The front-view camera is demonstrated in Figure~\ref{fig:app:task_showcase}; the wrist camera angle is shown in Figure~\ref{fig:qualitative-threepiece} (right); the extra Threading side-view camera is shown in Figure~\ref{fig:qualitative-threading} (right).

\subsection{Real-World Tasks}
\label{sec:app:tasks_real}

\begin{figure}[ht]
     \centering
     \begin{subfigure}[b]{0.49\textwidth}
         \centering
         \includegraphics[width=\textwidth]{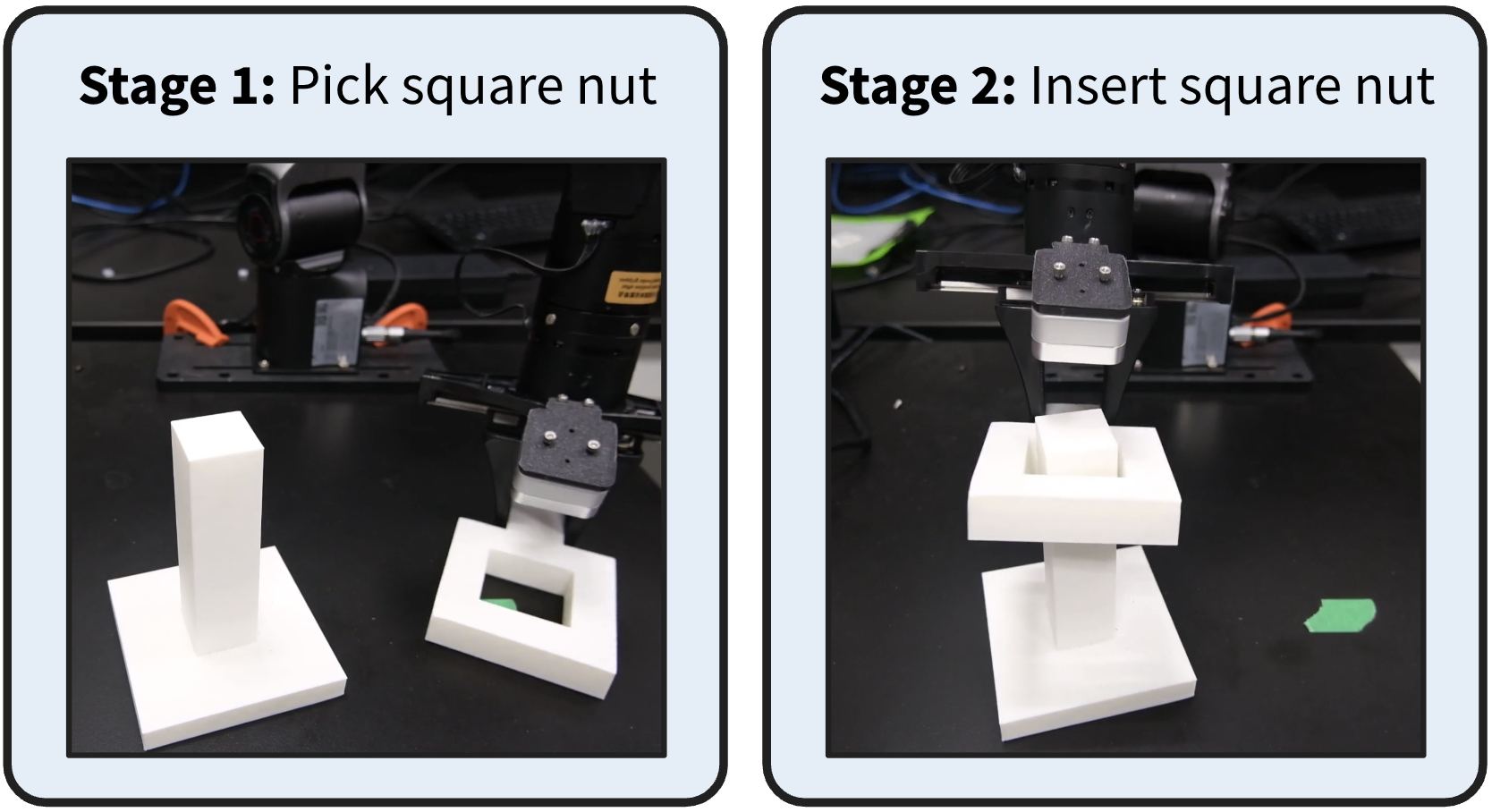}
         \caption{Square (2 stages)}
         \label{fig:app:env_square_real}
     \end{subfigure}
     \hfill
     \begin{subfigure}[b]{0.49\textwidth}
         \centering
         \includegraphics[width=\textwidth]{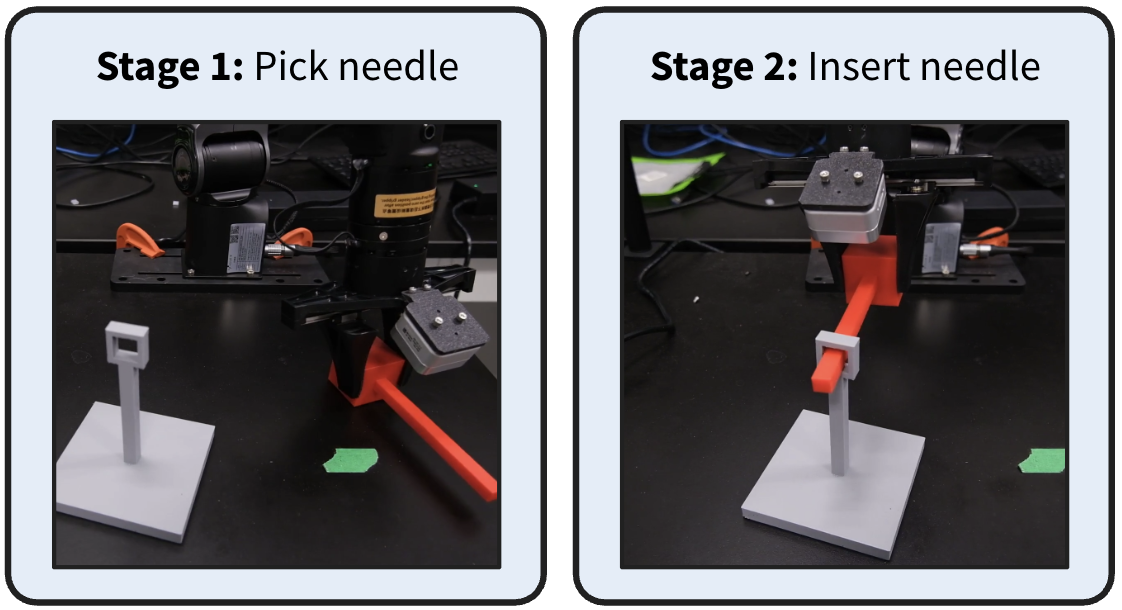}
         \caption{Threading (2 stages)}
         \label{fig:app:env_threading_real}
     \end{subfigure}
     \caption{Two real-world tasks.}
     \label{fig:app:task_showcase_real}
\end{figure}

\textit{Square} (first two stages of \textit{Nut Assembly}) and \textit{Threading} are selected for real-world evaluations (Fig.~\ref{fig:app:task_showcase_real}). We remove the legs of the ring stand in \textit{Threading} for 3D-printing compatibility. We also use a limited initialization range due to the size and controllability of the robot arm. The specific ranges are:

\textbf{Square}
\begin{itemize}
    \item Square nut: $x\in(-0.05, 0)$, $y\in(0.05, 0.1)$, $z_{\text{rot}}\in(-30^{\circ}, 30^{\circ})$
    \item Square peg: $x\in(-0.05, 0)$, $y\in(-0.1, -0.05)$, $z_{\text{rot}}\in(-30^{\circ}, 30^{\circ})$
\end{itemize}

\textbf{Threading}
\begin{itemize}
    \item Needle: $x\in(-0.05, 0)$, $y\in(0.05, 0.1)$, $z_{\text{rot}}\in(-30^{\circ}, 30^{\circ})$
    \item Ring stand: $x\in(-0.025, 0.025)$, $y\in(-0.1, -0.05)$, $z_{\text{rot}}\in(-15^{\circ}, 15^{\circ})$
\end{itemize}

\textbf{Observations.} For simulation training, we use the same observation space as our simulation task setup. We do not include the third side-viewing camera in \textit{Threading}, as the original two cameras suffice in this initialization setting. The robot joint configuration input also becomes 6-dimensional to accommodate the 6-DOF arm used in our evaluation.

\clearpage
\section{Reproducibility}
\label{app:reprod}

\subsection{Pseudocode}

\begin{algorithm}[H]
\begin{footnotesize} 
  \caption{\AlgoName{} Deployment Pseudocode} 
  \label{alg:policy}  
  \begin{algorithmic}[1] 
    \Procedure{\AlgoName{}}{}

        \State $o\gets \texttt{env.reset()}$ \Comment{Get initial observation}
        \For{$i:=1\rightarrow n$}
            \State $\langle R_i, \mathcal{I}_{\theta_i}, \pi_{\theta_i}, \mathcal{T}_{\theta_i} \rangle \gets \psi_{\theta_i}$ \Comment{Skill $\psi_i$}
            \State $p\gets \mathcal{I}_{\theta_i}(o)$ \Comment{Predict initiation pose}
            \State $\tau \gets \texttt{planToPose(\texttt{env}, $p$)}$ \Comment{Motion planning}
            \For{$a\in\tau$}
                \State $o\gets \texttt{env.step($a$)}$ \Comment{Execute motion action}
                \State $p'\gets \mathcal{I}_{\theta_i}(o)$
                \If {$\text{dis}(p,p')>\epsilon$} \Comment{Refined initiation prediction}
                    \State $p\gets p'$
                    \State $\tau \gets \texttt{planToPose(\texttt{env}, $p$)}$ \Comment{Replan trajectory}
                \EndIf 
            \EndFor
            \While{$\mathcal{T}_{\theta_i}(o) \neq 1$} \Comment{Until subgoal success}
                \State $a \gets \pi_{\theta_i}(o)$
                \State $o\gets \texttt{env.step($a$)}$ \Comment{Execute policy action}
            \EndWhile
        \EndFor
    \EndProcedure
\end{algorithmic}
\end{footnotesize}
\end{algorithm}

\subsection{Depiction of Fine-Tuning}

\begin{figure}[t]
     \centering
     \centering
     \includegraphics[width=0.95\textwidth]{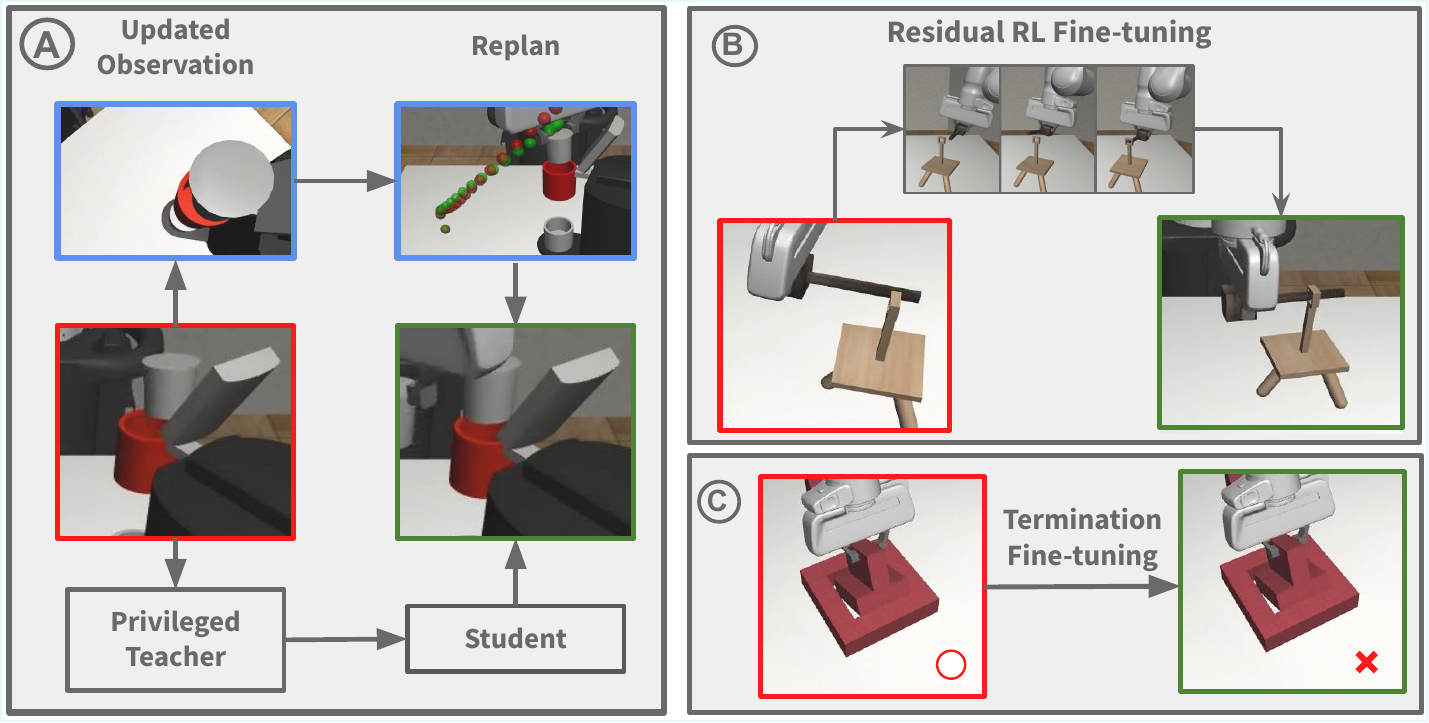}
        \caption{Depiction of how we fine-tune the three components. (A) The pose predictor based on image and robot state input is fine-tuned towards a teacher predictor with privileged object state information. During execution, it constantly updates its prediction when gathering new information about the target object and reroutes dynamically. (B) The skill policy is fine-tuned through residual reinforcement learning. (C) We purge the false-positive stage termination predictions that lower the completion rate of later stages.
        }
        \label{fig:fine-tune}
\end{figure}

Fig.~\ref{fig:fine-tune} illustrates the fine-tuning process of \AlgoName{} on the three components of an HSP.

\subsection{\rebuttal{}{Thresholds Used in \AlgoName{}}}

\textbf{Threshold for motion planner replanning (Sec.~\ref{sec:alg:init-cond})} We replan the motion planning trajectory when the pose distance (c.f. App.~\ref{app:sec:pose-distance}) between the newest predicted pose and the current pose target exceeds \textbf{0.05}.

\textbf{Threshold for termination rejection (Sec.~\ref{sec:alg:term-cond})} We reject terminations with a predicted task completion rate lower than \textbf{0.4}.

\subsection{Hyperparameters for Data Generation and Imitation Learning}

We follow the exact setup in \cite{garrett2024skillmimicgen}.

\subsection{Hyperparameters for Reinforcement Learning}
\label{app:rl}

\begin{table}[h]
\centering
    \caption{DrQ-v2 hyperparameters.}
    \label{tab:rl-hp}
    \begin{tabular}{lc}
        \toprule
        Network structure & CNN \\
        Learning rate  & 1e-4 \\
        Discount & 0.99 \\
        Batch size             & 256 \\
        $n$-step returns & 3 \\
        Action repeat & 1 \\
        Seed frames & 4000 \\
        Feature dim & 50  \\
        Hidden dim & 1024  \\
        Optimizer & Adam  \\
        Training steps & 2M \\
        \midrule
        Constraint coef. ($\alpha$) & 5.0 \\
        \bottomrule
    \end{tabular}
\end{table}

We use DrQ-v2~\citep{yarats2021drqv2} for skill fine-tuning (Sec.~\ref{sec:alg:policy}). The hyperparameters are shown in Tab.~\ref{tab:rl-hp}.

\subsection{Usage of Fine-tuning Methods}
\label{app:ft-impl}

\newcommand{\cmark}{\ding{51}}%
\newcommand{\xmark}{\ding{55}}%

\begin{table}[h]
\centering
    \caption{Usage of different fine-tuning methods in \AlgoName{}}
    \label{tab:ft-impl}
    \begin{tabular}{lcccc}
        \toprule
        Task & Stage & Pose distillation (Sec.~\ref{sec:alg:init-cond}) & Skill fine-tune (Sec.~\ref{sec:alg:policy}) & Termination fine-tune (Sec.~\ref{sec:alg:term-cond}) \\
        \midrule
        Coffee & 1 & \xmark & \xmark & \xmark \\
         & 2 & \xmark & \cmark & \xmark \\
        \midrule
        Threading & 1 & \xmark & \xmark & \xmark \\
         & 2 & \xmark & \cmark & \xmark \\
        \midrule 
        Nut Assembly & 1 & \cmark & \xmark & \xmark \\
         & 2 & \xmark & \cmark & \xmark \\
         & 3 & \cmark & \cmark & \xmark \\
         & 4 & \xmark & \xmark & \xmark \\
        \midrule 
        Three Piece & 1 & \xmark & \xmark & \cmark \\
         & 2 & \xmark & \cmark & \cmark \\
         & 3 & \xmark & \xmark & \cmark \\
         & 4 & \xmark & \cmark & \cmark \\
        \midrule 
        Coffee Prep. & 1 & \xmark & \xmark & \xmark \\
         & 2 & \xmark & \cmark & \xmark \\
         & 3 & \xmark & \cmark & \xmark \\
         & 4 & \xmark & \cmark & \xmark \\
         & 5 & \xmark & \cmark & \xmark \\
        \bottomrule
    \end{tabular}
\end{table}

By default, all stages in all tasks use first-iteration pose distillation and real-time replanning (Sec.~\ref{sec:alg:init-cond}). The usage of second-iteration pose distillation and skill/termination fine-tuning is listed in Tab.~\ref{tab:ft-impl}.

\subsection{End-to-End Distillation}

We add Gaussian noise $\mathcal{N}(0,\sigma^2)$ with $\sigma=0.01$ during trajectory generation. For each task, we generate 3000 successful trajectories. We use the same hyperparameters as skill imitation for IL.
\clearpage
\section{Additional Experiments}
\label{app:add-results}


\subsection{Case Study on the Effectiveness of Replanning in \emph{Nut Assembly}}
\label{app:ablation-replan}


\begin{figure}[h]
    \centering
    \includegraphics[width=0.34\linewidth]{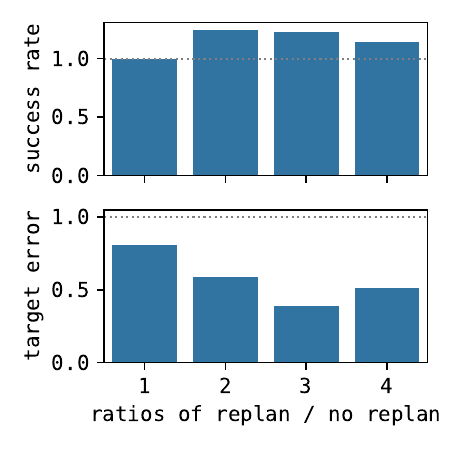}
    \caption{Ablation of real-time replanning at every stage of \emph{Nut Assembly}. The top figure shows the per-stage success rate, and the bottom shows the pose target error. The numbers are ratios of applying replanning versus not applying.}
    \label{fig:ablation-replan}
\end{figure}

To illustrate the effectiveness of real-time replanning (Sec.~\ref{sec:alg:init-cond}), we plot the reduction in prediction error and improvement in task success rate with replanning in all 4 stages in \textit{Nut Assembly}. The results in Fig.~\ref{fig:ablation-replan} show that replanning significantly reduces the target prediction error in all stages, and in turn, improves the subsequent skill success rates.

\subsection{Learned Termination Predictor}

\textbf{Learned termination predictors have limited impact on task completion.} We evaluate \AlgoName{} with a learned termination predictor that determines termination with partial observations. As shown in Table~\ref{tab:learned-term}, 
using a learned termination predictor only introduces minor drops in success rates in \emph{Threading} and \emph{Three Piece}, while maintaining its performance in the other tasks.

\begin{table}[h]
\vspace{-3mm}
\centering
\begin{footnotesize}
\begin{tabular}{|l|rrrr|}
      \toprule
\textbf{Success Rate (\%)} & \emph{Nut Assembly} & \emph{Threading} & \emph{Three Piece} & \emph{Coffee} \\
      \midrule
Oracle Termination & 85.80 & 82.20 & 80.40 & 93.81 \\
Learned Termination & 84.60 & 79.20 & 73.80 & 92.60 \\
    \bottomrule
  \end{tabular}
    \caption{
    \AlgoName{}'s learned termination predictor, which unlike the Oracle Termination does not have access to the state, performs well albeit slightly worse when compared to the Oracle Termination upper bound.
    }
    \label{tab:learned-term}
\vspace{-3mm}
\end{footnotesize}
\end{table}

\subsection{Ablation on Distillation}

\begin{figure}[h]
    \centering
    \includegraphics[width=1.0\linewidth]{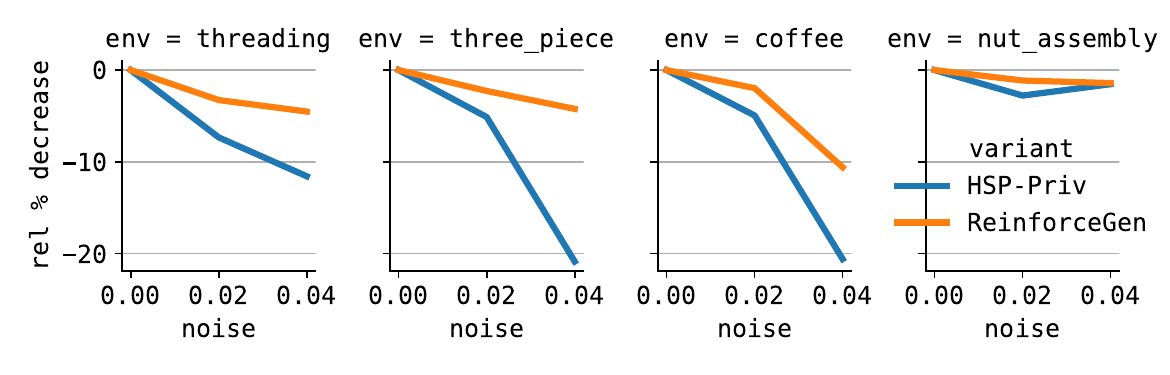}
    \caption{Comparing the resistance to action noise between HSP-Priv and \AlgoName{}. For most tasks, \AlgoName{} has a much higher tolerance to action noise.}
    \label{fig:distill-noise-succ}
\end{figure}

In Fig.~\ref{fig:distill-noise-succ}, we add artificial noise to the actions and plot the relative performance decreases against the noise scale. In all tasks, \AlgoName{} appears to be more resistant to action noise and trajectory deviations, matching our assertions in Sec.~\ref{sec:alg:distill}.

\subsection{Real-World Evaluation on \textit{Threading}}
\label{sec:app:real-results-threading}

We also evaluated \AlgoName{} on a real-world version of \textit{Threading} (see Sec.~\ref{sec:app:tasks_real}). We collected only 1 tele-op demonstration and generated 100 demonstrations for imitation learning to test the limit of \AlgoName{}. We also only fine-tuned the second stage skill policy since it is the most challenging part of the task. The results are shown in Tab.~\ref{tab:sim2real-threading}.

\begin{table}[h]
\vspace{-3mm}
\begin{footnotesize}
\centering
\begin{tabular}{|l|rrr|}
      \toprule
\textbf{Success Rate (\%)} & \emph{Simulation Success.} & \emph{Real Success.} & \emph{Sim-to-Real Success.} \\
      \midrule
HSP & 20.00 & 5.00 & 25.00 \\
ReinforceGen & \textbf{75.00} & \textbf{35.00} & \textbf{46.67} \\
      \bottomrule
\end{tabular}
\caption{Real-world evaluation results on \textit{Threading}. Success rates are averaged over 20 rollouts. Sim-to-real transfer success rates are calculated by dividing real-world success rate by simulation success rate.
}
\label{tab:sim2real-threading}
\end{footnotesize}
\vspace{-5mm}
\end{table}

Consistent with the conclusion drawn in the \textit{Square} experiments, The \AlgoName{} agent not only shows a significantly higher success rate than HSP on a real robot, but also is more effective in handling the sim-to-real gap.
\clearpage
\section{\rebuttal{}{Additional Details}}
\label{app:details}

\subsection{\rebuttal{}{Pose Noise in Section~\ref{sec:alg:init-cond}}}
\label{app:details:pose-noise}

In Figure~\ref{fig:error-succ}, we add artificial noise to initiation poses to demonstrate their relationship with skill completion rates. The noise is added with the following procedure. Let the original pose be $\boldsymbol{P}:=\boldsymbol{p}\oplus\boldsymbol{r}$, where $\boldsymbol{p}$ is a 3-dimensional position vector in meters, $\boldsymbol{r}$ is a 3-dimensional rotation vector in the axis-angle form. A noisy version of $\boldsymbol{P}$ with a scale $\sigma$ is defined as:

\begin{equation}
    \tilde{\boldsymbol{P}}\leftarrow\boldsymbol{P}+\mathcal{N}(\boldsymbol{0}, \sigma^2\boldsymbol{I}_6).
\end{equation}

\subsection{\rebuttal{}{Pose Distance Metric}}
\label{app:sec:pose-distance}

We use the following metric to compute the distance between two poses throughout our implementation. Let $\mathbf{P_1}:=(\mathbf{p_1}, \mathbf{q_1})$, $\mathbf{P_2}:=(\mathbf{p_2}, \mathbf{q_2})$ be the two poses to compare, $\mathbf{p_1}, \mathbf{p_2}$ the Euclidean positions, and $\mathbf{q_1}, \mathbf{q_2}$ the quaternion-form rotations.

\begin{align}
    d^\text{pos}(\mathbf{P_1}, \mathbf{P_2})&:=\sqrt{(\mathbf{p_1}-\mathbf{p_2})^\top(\mathbf{p_1}-\mathbf{p_2})}\\
    d^\text{rot}(\mathbf{P_1}, \mathbf{P_2})&:=\frac{1}{\pi}\min\{\cos^{-1}(\mathbf{q_1}^\top\mathbf{q_2}), \cos^{-1}(-\mathbf{q_1}^\top\mathbf{q_2})\}\\
    d^\text{pose}(\mathbf{P_1}, \mathbf{P_2})&:=d^\text{pos}(\mathbf{P_1}, \mathbf{P_2})+ d^\text{rot}(\mathbf{P_1}, \mathbf{P_2})
\end{align}
\clearpage
\section{Real-World Setup}
\label{app:real}

We provide additional details on our real-world evaluations. See Sec.~\ref{sec:app:tasks_real} for the details of the real-world tasks. See Sec.~\ref{sec:app:real-results-threading} for evaluation results on \textit{Threading}.

\textbf{Robot.} We use an AgileX PiPER 6-DOF arm for all our real-world evaluations. The arm is mounted at a 90-degree angle, facing the negative $y$-axis.

\textbf{Pose estimation.} We use a fixed RealSense D435i camera to capture the scene and FoundationPose~\citep{foundationposewen2024} to estimate the object poses. We do pose estimation once at the beginning of an episode.

\textbf{Demonstrations.} We use another PiPER teacher arm to tele-op the main arm and collect demonstrations in the real world. We then replay the joint position trajectory of the tele-op demonstrations in a MuJoCo simulator and convert them into a simulation-based demonstration dataset. For \textit{Square}, we collected 10 tele-op demonstrations and generated a 1,000-demonstration dataset. For \textit{Threading}, we collected 1 tele-op demonstration and generated a 100-demonstration dataset. We use the same setup as our simulation evaluations to train a BC policy.

\textbf{Fine-tuning.} Since pose estimation is required in our workflow, we do not need to do object-state-agnostic initiation pose prediction. We therefore use the privileged pose predictor as in HSP-Priv and omit the initiation pose prediction fine-tuning process. We use the same setup as in our simulation experiments for skill and termination fine-tuning. For \textit{Threading}, we omit termination fine-tuning and only fine-tune the second-stage skill policy.

\begin{figure}[ht]
     \centering
     \begin{subfigure}[b]{0.49\textwidth}
         \centering
         \includegraphics[width=\textwidth]{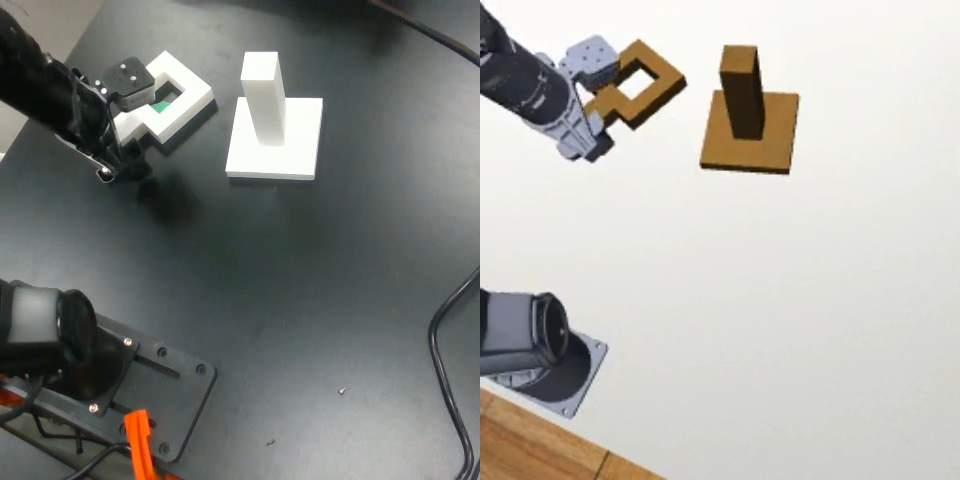}
         \caption{Square}
         \label{fig:app:env_square_sim2real}
     \end{subfigure}
     \hfill
     \begin{subfigure}[b]{0.49\textwidth}
         \centering
         \includegraphics[width=\textwidth]{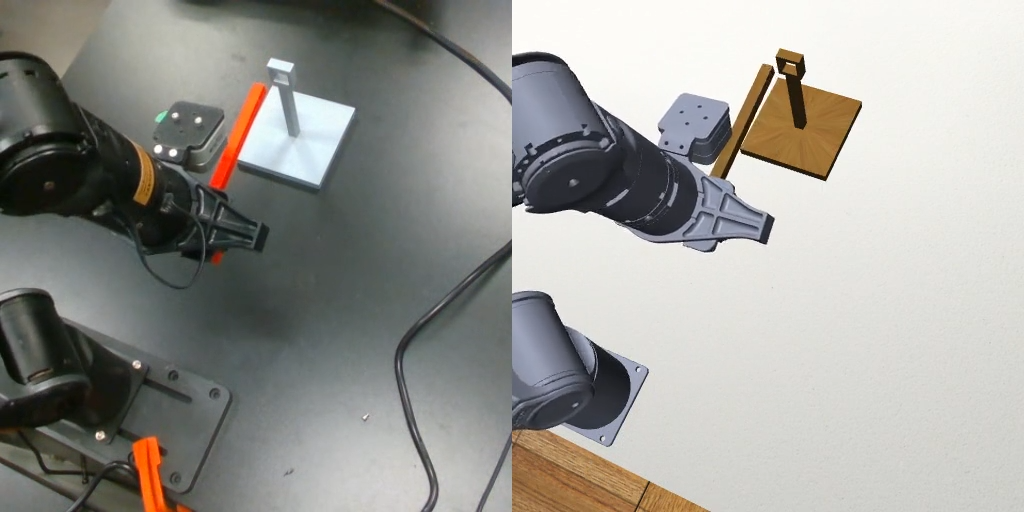}
         \caption{Threading}
         \label{fig:app:env_threading_sim2real}
     \end{subfigure}
     \caption{Sim-to-real deployment.}
     \label{fig:app:sim_to_real}
\end{figure}

\textbf{Deployment.} We use the following procedure to deploy our simulation-based policy on a real robot. First, we use FoundationPose to estimate the poses of the objects in the real world. We then sync the estimated poses in our MuJoCo simulation and run the evaluated policy in simulation, while recording the joint position and gripper action trajectory. If the simulation is successful, we finally replay the trajectory in the real world by sending joint and gripper commands to the robot arm. See Fig.~\ref{fig:app:sim_to_real} for an illustration.

\end{document}